\pdfoutput=1

\documentclass[11pt]{article}

\usepackage[table, dvipsnames]{xcolor}

\usepackage[preprint]{acl}

\usepackage{times}
\usepackage{latexsym}
\usepackage{calc}
\usepackage{listings} 

\usepackage[T1]{fontenc}

\usepackage[utf8]{inputenc}

\usepackage{microtype}

\usepackage{inconsolata}

\usepackage{graphicx}
\usepackage{xcolor}         
\usepackage{graphicx}
\usepackage{longtable} 
\usepackage{booktabs} 
\usepackage{array} 
\usepackage{lipsum} 
\usepackage{multirow}
\usepackage{makecell}
\usepackage{pifont}
\newcommand{\xmark}{\ding{55}}%
\usepackage{tabularx}
\usepackage{amssymb}
\usepackage[table]{xcolor} 
\usepackage{colortbl}    
\usepackage{tikz}
\usepackage[most]{tcolorbox}

\usepackage{setspace}
\usepackage{adjustbox}
\definecolor{placeholdercolor}{HTML}{DB5A6B} 

\usepackage{enumitem} 
\usepackage{amsmath} 

\usepackage{xspace}
\newcommand{\methodname}{\textcolor{black}{\textsc{BehaviorBench}}\xspace}

\title{BehaviorSFT: Behavioral Token Conditioning for Clinical Agents \\ Across the Proactivity Spectrum}

\author{%
  Yubin Kim\textsuperscript{1},
  Zhiyuan Hu\textsuperscript{1,2},
  Hyewon Jeong\textsuperscript{1},
  Eugene W Park\textsuperscript{1},
  Shuyue Stella Li\textsuperscript{3},\\ 
  Chanwoo Park\textsuperscript{1}, 
  Shiyun Xiong\textsuperscript{2},
  MingYu Lu\textsuperscript{3},
  Hyeonhoon Lee\textsuperscript{4},
  Xin Liu\textsuperscript{5},\\ 
  Daniel McDuff\textsuperscript{5},
  Cynthia Breazeal\textsuperscript{1},
  Samir Tulebaev\textsuperscript{6},
  Hae Won Park\textsuperscript{1}\\ 
  \\ 
  \textsuperscript{1}Massachusetts Institute of Technology,
  \textsuperscript{2}National University of Singapore,
  \textsuperscript{3}University of Washington, \\ 
  \textsuperscript{4}Seoul National University Hospital,
  \textsuperscript{5}Google,
  \textsuperscript{6}Mass General Brigham  \\
  \\ 
  \small{
    \textbf{Correspondence:} \href{mailto:ybkim95@mit.edu}{ybkim95@mit.edu}
  }
}

\begin{document}
\maketitle
\begin{abstract}
Large Language Models (LLMs) as clinical agents require careful behavioral adaptation. While adept at reactive tasks (e.g., diagnosis reasoning), LLMs often struggle with proactive engagement, like unprompted identification of critical missing information or risks. We introduce \textbf{BehaviorBench}, a comprehensive dataset to evaluate agent behaviors across a clinical assistance spectrum, ranging from reactive query responses to proactive interventions (e.g., clarifying ambiguities, flagging overlooked critical data). Our BehaviorBench experiments reveal LLMs' inconsistent proactivity. To address this, we propose \textbf{BehaviorSFT}, a novel training strategy using  behavioral tokens to explicitly condition LLMs for dynamic behavioral selection along this spectrum. BehaviorSFT boosts performance, achieving up to 97.3\% overall Macro F1 on BehaviorBench and improving proactive task scores (e.g., from 95.0\% to 96.5\% for \texttt{Qwen2.5-7B-Ins}). Crucially, blind clinician evaluations confirmed BehaviorSFT-trained agents exhibit more realistic clinical behavior, striking a superior balance between helpful proactivity (e.g., timely, relevant suggestions) and necessary restraint (e.g., avoiding over-intervention) versus standard fine-tuning or explicit instructed agents.\footnote{Project Page: \\ \url{https://behavior-adaptation.github.io/}}
\end{abstract}

\section{Introduction}
As Large Language Models (LLMs) transition from experimental systems to deployed agents in clinical environments, a critical question emerges: ``\textit{when} should these systems act \emph{reactively} or \emph{proactively} \citep{fauscette2024agentic}?.'' Unlike general-purpose AI agents, healthcare agents can operate in high-stakes environments where both action and inaction carry significant consequences \citep{kim2025medical}.
We define \textit{reactive} behaviors as those where the agent responds only to explicit queries with precisely the information requested, while \textit{proactive} behaviors involve volunteering additional information, raising concerns, or suggesting actions beyond what was directly solicited. Importantly, proactivity in clinical contexts extends beyond merely asking clarifying questions, a common, but limited, focus in existing NLP research \citep{li2024mediq, hu2024uncertainty}. While question-asking represents one dimension of proactivity, our work encompasses a broader spectrum - including unsolicited intervention, critical evaluation, and recommendation. These behaviors align closely with the "Appraisal" phase of Evidence-Based Medicine (EBM) \citep{denby2008evidence}, where clinicians actively assess available information, identify information gaps, and determine appropriate next steps. An agent that remains strictly reactive may fail to raise an alert when problems are observed with critical lab values or medication contraindications \citep{walter2021clinical, wright2018reduced}, potentially compromising patient safety \citep{mccoy2014clinical}. In contrast, an excessively proactive system that frequently interrupts with unsolicited recommendations risks contributing to alert fatigue, interruption of workflow, and potential rejection by healthcare professionals \citep{sutton2020overview}.
This trade-off between reactive and proactive behaviors forms the core challenge addressed in this paper. The appropriate balance between these modalities varies dramatically based on clinical context, urgency, risk levels, and the specific healthcare roles being augmented, demanding adaptive behavior policy rather than a fixed mode, especially as systems achieve higher levels of autonomy (Figure \ref{fig:behavioral_adaptation}).
\begin{figure*}[t!]
\centering
\includegraphics[width=\textwidth]{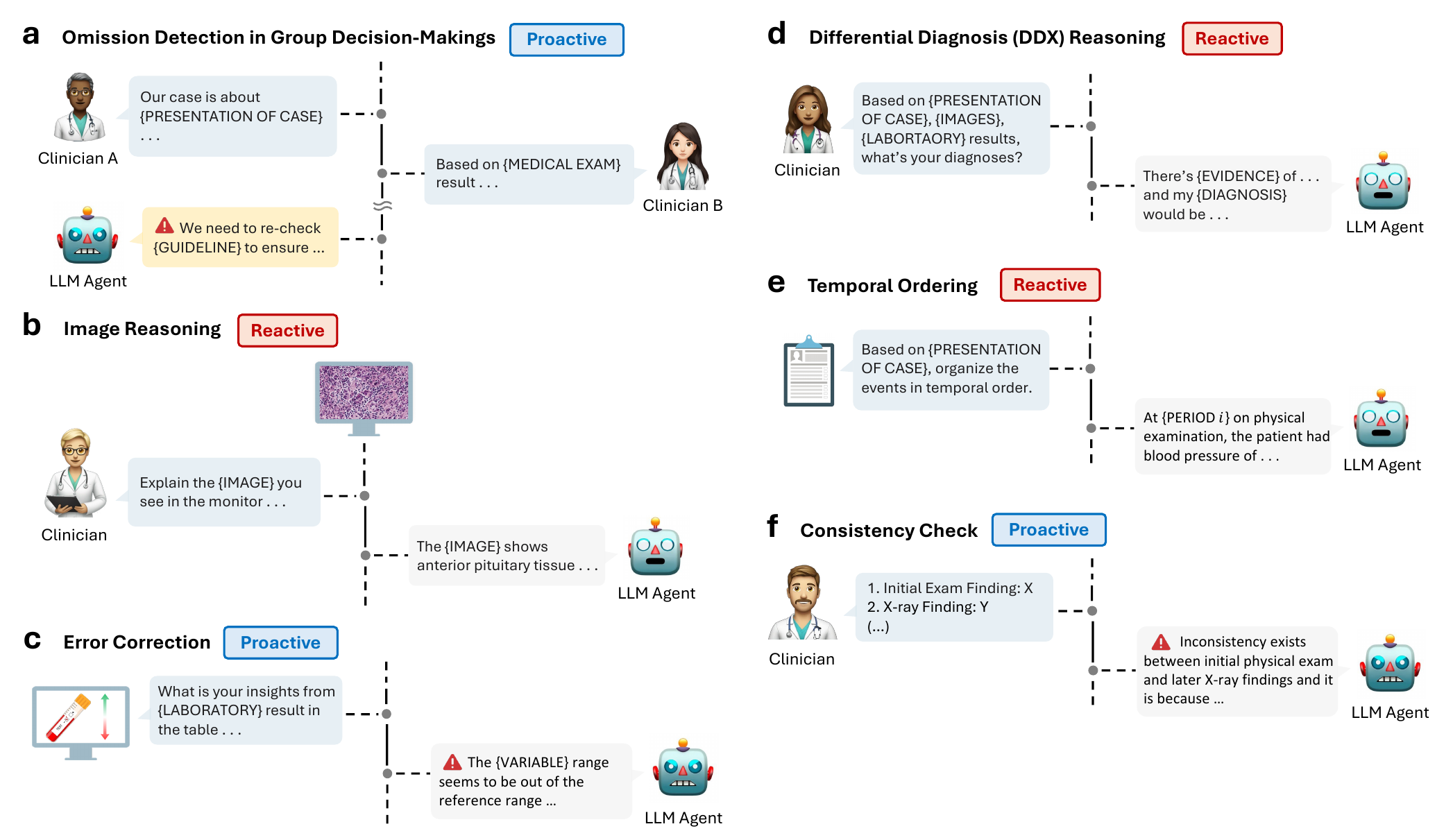}
\caption{\textbf{Six representative tasks from \methodname, showcasing the spectrum of agent behaviors in clinical settings.} The figure illustrates (a-c, f) proactive tasks where the LLM agent identifies issues or offers insights without direct prompting, and (b, d, e) reactive tasks responding to explicit clinician queries.}
\label{fig:tasks} 
\end{figure*}

To systematically discuss how an agent's reactive and proactive stance should adapt with its increasing capabilities, we adapt the SAE Levels of Driving Automation \cite{SAEJ3016_202104} into a six-level taxonomy for healthcare AI agent autonomy. This framework detailed in Table~\ref{tab:autonomy_taxonomy_detailed} in Appendix helps to illustrate a key principle: as an AI agent ascends these autonomy levels, its capacity and responsibility to engage in sophisticated proactive behaviors, rather than merely reactive ones, become increasingly critical.






The autonomy level taxonomy highlights that effective healthcare AI, particularly for achieving \textbf{Level 3 (Conditional Proactive Assistance) and above}, must move beyond simple reactive responses (Levels 1-2). As AI autonomy increases, the nature of clinician responsibility evolves, shifting from direct task execution to supervision, validation of AI-driven insights, and management of exceptions. Our work, therefore, focuses on enabling AI agents to learn and exhibit the adapted spectrum of reactive and proactive behaviors crucial for safe and effective operation at these higher levels of conditional and collaborative automation. \methodname{} is designed to evaluate these capabilities across this spectrum, and \textbf{BehaviorSFT} aims to train agents to achieve this behavioral adaptability, particularly for robust performance at Levels 2 and 3, with an eye towards future capabilities at Level 4.


Effectively adapting \textit{which} of these behaviors is appropriate, and \textit{when}, is essential for clinical AI systems that can safely operate at increasing levels of autonomy.
In this work, we ask \textit{what proactivity means for healthcare AI and how we build systems that are appropriately behaving?}
To this end, we propose a novel six-level taxonomy for healthcare AI autonomy that maps progression from human-controlled to autonomous operation.
We trace the evolution from early reactive systems \citep{tu2024towards,han2023medalpaca} to more recent developments like MediQ \citep{li2024mediq} and AIME \citep{mcduff2025towards, tu2024towards}, which incorporate proactive elements while demonstrating the critical interplay between proactivity and urgency.
Our benchmark was curated from real medical cases sourced from New England Journal of Medicine (NEJM) clinical case reports \citep{brinkmann2024building}. We employed a LLM (\texttt{Gemini-2.5 Flash}) to meticulously ground these cases in their factual details and then reformat them into multi-turn, multi-clinician-patient conversational scenarios, integrating multi-modal inputs such as text, images, and tabular data. Indeed, we propose this LLM-assisted methodology for converting existing static clinical datasets into rich, reactive-proactive benchmark scenarios as a key contribution of our work.
Additionally, we present a novel training methodology, \textbf{BehaviorSFT}, which employs explicit behavioral tokens to condition LLM responses along the reactive-proactive spectrum. Our approach demonstrates significant improvements, achieving up to 97.3\% overall Macro F1 on BehaviorBench (compared to 96.7\% for general SFT) with particularly notable gains in proactive tasks (from 95.0\% to 96.5\%). The primary contributions are:

\begin{enumerate}
\item We introduce \methodname, an evaluation dataset that assesses LLM capabilities across both reactive and proactive tasks in healthcare contexts.
\item We provide detailed analysis of recent LLMs' performance on \methodname, revealing significant variability in contextual awareness and appropriate behavioral adaptation.

\item We propose BehaviorSFT, a new fine-tuning strategy that leverages behavioral tokens to guide LLMs in dynamically adapting their responses along the reactive-proactive tasks.
\end{enumerate}

\begin{table*}[t!]
\centering
\footnotesize
\caption{\textbf{Comparison of Public Medical Benchmarks.} Modality codes: \textit{t}=text, \textit{i}=image, \textit{b}=tabular/structured data. \checkmark \ indicates that the benchmark natively supports the evaluation dimension; \xmark \ indicates it does not.}
\label{tab:medical_benchmarks}
\begin{tabular}{lcccccc}
\toprule
\textbf{Benchmark} & \makecell{\textbf{Size}} & \textbf{Modality} & \makecell{\textbf{Behavior}\\\textbf{Evaluation}} & \makecell{\textbf{Sequential}\\\textbf{Eval.}} & \makecell{\textbf{Dialogue}\\\textbf{Interaction}} & \makecell{\textbf{Multiple}\\\textbf{Roles}} \\
\midrule
MedQA \citep{jin2021disease}           & 1{,}273   & \textit{t}             & \xmark & \xmark & \xmark & \xmark \\


MedMCQA \citep{pal2022medmcqa}                & 6{,}100 & \textit{t}             & \xmark & \xmark & \xmark & \xmark \\

MultiMedQA \citep{singhal2023large}     & 13{,}115 & \textit{t}             & \xmark & \xmark & \xmark & \xmark \\

MediQ \citep{li2024mediq}         & 1{,}273  & \textit{t}             & \xmark & \checkmark & \checkmark & \checkmark \\

MediQ-AskDocs \citep{li2025aligning}         & 17{,}000  & \textit{t}             & \xmark & \checkmark & \checkmark & \checkmark \\

ClinicBench \citep{chen2024clinicalbench}            & 11{,}000  & \textit{t}             & \xmark & \xmark & \xmark & \xmark \\

MedChain \citep{liu2024medchain}           & 12,163  & \textit{t}+\textit{i}  &  \xmark  &  \checkmark  &  \checkmark  &  \checkmark  \\

MedAgentBench \citep{jiang2025medagentbench}         & 300       & \textit{t}+\textit{b}  & \xmark & \checkmark & \checkmark & \checkmark \\

HealthBench \citep{arora2025healthbench} & 5,000 & \textit{t} & \xmark & \xmark & \checkmark & \xmark \\
\midrule

\rowcolor{green!8} \textbf{\methodname (Ours)} & 142,496 & \textit{\textbf{t}}+\textit{\textbf{i}}+\textit{\textbf{b}} & \textbf{\checkmark} & \textbf{\checkmark} & \textbf{\checkmark} & \textbf{\checkmark} \\
\bottomrule
\end{tabular}
\end{table*}

\section{\methodname}
\label{sec:behaviorbench}

We introduce \methodname, a novel dataset specifically designed to assess agent capabilities across the reactive-proactive tasks. Derived from real clinical cases, \methodname comprises of 6,876 real-world clinical case scenarios from which we derived a total of 142,496 tasks distributed across the 13 distinct task categories. This framework provides a more granular analysis of an agent's ability to discern context and modulate its behavior accordingly, moving beyond standard metrics, such as accuracy, that are solely based on reactive responses. Detailed dataset statistics can be found in the Appendix \ref{sec:dataset_statistics}.

To ensure that the generated tasks effectively probe clinical reasoning, we construct the dataset in a two-step process. First, we carefully prompt the LLM (see Appendix \ref{sec:prompt_template}) generating the tasks to use detailed summary from real-world clinical cases, including patient history, diagnostics, conversation snippets, and final diagnoses. This ensures that the questions, answers, and rationales reflect genuine clinical context instead of relying on pseudolabels generated without any realistic groundings. All draft tasks then underwent several back‑and‑forth revision cycles with two physicians, who reviewed any hallucinations and confirmed each scenario’s practical plausibility for N=10 cases. Then, to evaluate the agent's proactive capabilities, we augment the base scenarios by intentionally introducing subtle challenges, such as hypothetical scenarios with probable clinical errors, conflicting data points (e.g. modifying numerical values slightly between reports, or presenting exam findings seemingly at odds with imaging), and omitted information expected by clinical standards. The resulting reactive-proactive tasks are as follows:

\paragraph{Reactive Tasks} evaluates whether the agents can handle information when requested directly.
\begin{enumerate}
    \item \texttt{fact\_retrieval}: Finds specific facts mentioned in the text (e.g., ``What was the patient's initial temperature?").
    \item \texttt{timeline\_sequence}: Puts events in order using clear time references (e.g., tracing how lung exam findings changed between the initial presentation and Turn $N$, based on provided descriptions from those time points).
    \item \texttt{ddx\_reasoning}: Explains the reasoning for a possible diagnosis using only the evidence given (e.g., identifying findings prior to Turn $M$, such as specific X-ray descriptions and sputum results, that suggested bronchopneumonia over simple lobar pneumonia).
    \item \texttt{treatment\_decision}: Connects a doctor's thinking or action to the stated reason or data supporting it (e.g., evaluating a specific diagnostic leaning mentioned in Turn $K$ based only on the evidence explicitly available at that time, like sputum results).
\end{enumerate}

\paragraph{Balanced Tasks} are initiated by specific, provided information but demand a more significant cognitive step involving deeper thinking, such as multi-step inference, synthesis of multiple data points, or evaluating the impact of new information on existing understanding.

\begin{enumerate}
    \item \texttt{reasoning\_differential\_evolution}: Compares the patient's situation at two different times and explains how the doctor's assessment should change because of new information (e.g., asking how the list of possible diagnoses should shift from Timepoint A to Timepoint B considering newly available sputum culture results and vital signs).
    \item \texttt{integrity\_missing\_turn\_inference}: Figures out what was likely said in a missing part of a conversation based on what came before and after (e.g., ``Turn $N$ orders a test, Turn $N+M$ discusses the result. What likely happened in Turn $N+K$, where $0 < K < M$?'').
\end{enumerate}

\paragraph{Proactive Tasks} require the LLM to use higher-level thinking, and evaluation skills.

\begin{enumerate}
\item \texttt{predictive\_next\_action}: Forecasts the most appropriate subsequent clinical action by integrating the evolving patient case, current symptoms, medical history, and available diagnostic results.
\item \texttt{explicit\_error\_correction}: Identifies and rectifies explicitly stated errors in clinical narratives or proposed actions, providing justifications based on medical knowledge and case specifics (e.g., correcting drug suitability given a patient's allergy).
\item \texttt{omission\_detection}: Identifies significant omissions in the provided clinical information or documented actions, such as overlooked diagnostic tests or unaddressed critical symptoms that could impact patient care.
\item \texttt{standard\_of\_care}: Assesses whether documented clinical management, including diagnostic procedures and interventions, adheres to established medical guidelines and accepted best practices, often requiring external knowledge.
\item \texttt{interpretation\_conflict}: Discerns and reconciles nuanced or potentially conflicting interpretations of clinical findings from different sources (e.g., contrasting physical exam notes with radiology findings), articulating their clinical significance.
\item \texttt{data\_conflict\_resolution}: Identifies direct contradictions or inconsistencies between pieces of factual clinical data presented within a case (e.g., conflicting lab values over time) and proposes logical explanations.
\item \texttt{consistency\_check}: Evaluates the overall logical and clinical coherence of a case narrative or specific information, identifying elements that are incongruous or implausible (e.g., assessing if a patient's reported progression aligns with a given diagnosis).
\end{enumerate}


\section{BehaviorSFT: Behavior Adaptation Training}
\label{sec:behavior_adaptation_training}

To operationalize the concept of behavioral adaptation within healthcare LLM agents, we propose a targeted training strategy, Behavior-Conditioned Supervised Fine-Tuning (BehaviorSFT). This approach leverages our specialized \texttt{BehaviorBench} dataset (Section~\ref{sec:behaviorbench}) to explicitly teach LLMs to modulate their responses along the reactive-proactive spectrum based on inferred clinical context. This contrasts with standard SFT approaches, which typically optimize for task completion without explicit mechanisms to control the agent's level of initiative or caution, risking either unsafe passivity or disruptive over-intervention.

\subsection{Behavior Tokens}

\textbf{Rationale for Prefix Tokens:} We employ prefix behavior tokens (e.g., <reactive>, <proactive>) for several reasons. Placing the token at the beginning of the target sequence allows it to act as a direct control signal, conditioning the entire generation process on the desired behavioral mode from the outset. This explicitly trains the model to adopt the appropriate style, tone, and level of initiative as it generates the response. While one could consider predicting the token after some internal reasoning chain, our approach integrates this reasoning implicitly, i.e., the model learns to predict the correct initial token based on its understanding of the input context ($x$), as described in our Contextual Behavior Assessment capability (Section \ref{subsec:behaviorsft_revised}). This provides an end-to-end mechanism for context-aware, behaviorally adapted generation. Central to our approach is the introduction of special behavior tokens paired with the target response during training.
\begin{itemize}
    \item \texttt{<reactive>}: Signals the generation of a direct, concise response strictly adhering to the explicit query, avoiding unsolicited information or inferences.
    \item \texttt{<proactive>}: Signals a response that may include identifying implicit issues, volunteering relevant context or warnings, suggesting next steps, or applying external knowledge (e.g., standards of care) beyond the literal query.
\end{itemize}
These tokens act as control signals, learned by the model and conditioning the subsequent generation process. Alternative approaches exist, such as training a separate classifier to select the mode and then routing the input to specialized reactive or proactive models, or using inference-time techniques like thresholding logits associated with the behavior tokens for finer control. However, our BehaviorSFT approach offers a simpler, unified training process within a single model. Future work could explore hybrid methods or compare the efficacy of these different control paradigms.

\subsection{Training Data}
\label{subsec:data_prep_revised}
\texttt{BehaviorBench} serves as the crucial training ground for BehaviorSFT. Each instance within the benchmark's training split is meticulously annotated with the desired target behavior token based on the task's nature and the underlying clinical scenario's demands:

\begin{enumerate}
    \item \textbf{Reactive Annotation (\texttt{<reactive>}):} Applied to tasks demanding factual recall, direct sequencing, or simple reasoning strictly from provided data (e.g., \texttt{fact\_retrieval}, \texttt{timeline\_sequence}).
    \item \textbf{Proactive Annotation (\texttt{<proactive>}):} Applied to tasks necessitating critical assessment, error/omission detection, consistency checking, or prediction based on clinical standards (e.g., \texttt{consistency\_check}, \texttt{standard\_of\_care}, \texttt{predictive\_next\_action}).
    \item \textbf{Contextual Annotation for Balanced Tasks:} Instances from balanced tasks (e.g., \texttt{reasoning\_differential\_evolution}) are annotated based on whether the specific context warrants simple reporting (\texttt{<reactive>}) or highlighting significant changes/implications (\texttt{<proactive>}).
\end{enumerate}

Each annotated instance is then structured for auto-regressive SFT, pairing the input context/query with a target sequence beginning with the assigned behavior token, followed by an ideal response exemplifying that behavior.

    Example 1 (Reactive Task):
    \begin{lstlisting}
Input:  Context: [Note excerpt: Vitals stable.]
Query: Latest vitals?
Target: <reactive> BP 120/80, HR 75, Temp 37.0C, RR 16.
    \end{lstlisting}

    Example 2 (Proactive Task):
    \begin{lstlisting}
Input:  Context: [Chart: Rx Drug A. Allergy list: Drug A.]
Query: Confirm med list okay?
Target: <proactive> Warning: Drug A prescribed but patient is allergic. Review immediately.
    \end{lstlisting}

This structured data format explicitly teaches the model the association between clinical scenarios, appropriate behavioral modes (reactive/proactive), and corresponding linguistic outputs.

\subsection{Training Procedure: BehaviorSFT}
\label{subsec:behaviorsft_revised}
Starting with a pre-trained foundation LLM, we perform SFT using the behavior-annotated \texttt{BehaviorBench} training data. The objective is the standard causal language modeling loss, minimizing the negative log-likelihood of the target sequence $y = (y_1, ..., y_T)$, where $y_1 \in \{\texttt{<reactive>}, \texttt{<proactive>}\}$:
\begin{equation}
    \mathcal{L}_{BehaviorSFT} = - \sum_{i=1}^{T} \log P(y_i | y_{<i}, x; \theta)
    \label{eq:sft_loss_revised}
\end{equation}
Here, $x$ is the input context/query, $y_{<i}$ are the preceding target tokens, and $\theta$ represents the model parameters.

Through this process, the model learns the crucial, intertwined capabilities:
\begin{enumerate}
    \item \textbf{Contextual Behavior Assessment:} Implicitly analyzing the input $x$ to determine the likelihood that a proactive or reactive stance is warranted, influencing the prediction of the initial token $y_1$.
    \item \textbf{Behavior-Conditioned Generation:} Generating subsequent tokens $y_{2:T}$ in a manner consistent with the generated or given behavior token $y_1$, adopting the appropriate style, tone, and level of detail or intervention.
\end{enumerate}



\begin{table*}[t!]
\centering
\caption{\textbf{Performance on \methodname.} We report Macro F1-scores (\%) across three task categories. Best result per task is highlighted in \textbf{bold}.The \textbf{Ensemble} column reports baseline performance by majority voting across three commercial closed-source models (\texttt{Gemini-2.5-pro}, \texttt{OpenAI-o1}, \texttt{DeepSeek-R1}). `ZS' = Zero-Shot, `FS (k=3)' = Few-Shot (3 examples), `CoT' = Chain-of-Thought, `Explicit Instr.' = ZS with explicit reactive/proactive instruction, `Gen. SFT' = Standard Supervised Fine-Tuning (SFT), `BehaviorSFT' = Our proposed fine-tuning method.}
\label{tab:main_results1}
\resizebox{\textwidth}{!}{%
  \begin{tabular}{@{}lccccccccc@{}} 
    \toprule
    \textbf{Category} & \textbf{Task} 
      & \multicolumn{3}{c}{\textbf{Ensemble}} 
      & \multicolumn{2}{c}{\textbf{Qwen2.5-7B-Ins}} 
      & \multicolumn{2}{c}{\textbf{Llama3.1-8B-Ins}} \\
    \cmidrule(lr){3-5} \cmidrule(lr){6-7} \cmidrule(lr){8-9}
    & & ZS & FS (k=3) & ZS + Explicit Instr. 
      & Gen. SFT & \cellcolor{green!15}BehaviorSFT 
      & Gen. SFT & \cellcolor{green!15}BehaviorSFT \\
    \midrule
    \multirow{4}{*}{\rotatebox[origin=c]{90}{\textbf{Reactive}}}
    & fact\_retrieval          & 100.0 & 100.0 &    100.0      & 100.0 & 100.0 & 100.0 & 100.0 \\
    & timeline\_sequence       & 100.0 & 100.0  &   100.0       & 100.0 & 100.0 & 100.0 & 100.0 \\
    & ddx\_reasoning           & 96.2 & 96.6 &    96.6     &  96.1 &  96.1 & 94.2 & 92.7 \\
    & treatment\_decision      & 94.8 & 95.3 &    95.3     & 100.0 &  98.4 & 98.4 & 98.7 \\
    \cmidrule(lr){2-9}
    & \textit{Avg.}            & 98.2 & 98.2 &     98.2    & \textbf{98.6} &  \textbf{98.6} & 97.8 & 97.2 \\
    \midrule

    \multirow{2}{*}{\rotatebox[origin=c]{90}{\textbf{Balanced}}}
    & reasoning\_diff\_evolution    & 98.6 & 98.6 &   98.6      & 100.0 & 100.0 & 100.0 & 100.0 \\    
    & integrity\_missing\_turn      & 100.0 & 100.0 &    100.0      & 100.0 &  100.0 & 96.4 & 100.0 \\
    \cmidrule(lr){2-9}
    & \textit{Avg.}                   & 97.2 & 97.6 &    97.6     & \textbf{100.0} &  99.2 & 98.5  & \textbf{100.0} \\
    \midrule

    \multirow{7}{*}{\rotatebox[origin=c]{90}{\textbf{Proactive}}}
    & consistency\_check              & 94.3 & 100.0 &  94.3       & 100.0 & 100.0 & 100.0 & 100.0 \\
    & data\_conflict\_resolution     & 97.2&   97.2    &   97.2       &  99.3 & 98.6 & 99.2 & 98.6 \\
    & interpretation\_conflict        & 98.5 & 96.5 &  96.5       &  96.6 & 96.6 & 98.5 & 98.6 \\
    & standard\_of\_care              & 93.4 & 95.3  &    93.7     &  94.8 & 93.3  & 91.5 & 88.4 \\
    & omission\_detection              & 89.5     & 92.4      &    89.3     &  88.5 & 95.1  & 90.0 & 93.2 \\
    & explicit\_error\_correction      & 96.3 &  97.5 &   96.4      &  98.3 & 99.2 & 98.4 & 97.2 \\
    & predictive\_next\_action        & 82.5&  83.0&  82.3       &  84.8 & 91.7  &77.0  & 83.4 \\
    \cmidrule(lr){2-9}
    & \textit{Avg.}                    & 94.3 & 95.1 &    94.0     &  95.0 & \textbf{96.5}  & 94.2 & 94.7 \\
    \midrule

    \multicolumn{2}{@{}l}{\textbf{Avg.}} & 95.4 & 96.0&   95.3     & 96.7 & \textbf{97.3} &95.8  & 96.1 \\

    \bottomrule
  \end{tabular}%
}
\end{table*}

\begin{table*}[t!]
\centering
\caption{Macro F1‐scores of prompting methods on behavior classification.  
Method abbreviations: \textbf{BT} = Behavior token, \textbf{BC}  = Behavior chain‐of‐thought, \textbf{OC}  = Option CoT, \textbf{OP}  = Option.  
Class abbreviations:  
\textbf{Five-class} (BA = balanced; H\_PR = highly\_proactive; H\_RE = highly\_reactive; P\_PR = primarily\_proactive; P\_RE = primarily\_reactive),  
\textbf{Binary} (PR = proactive; N\_PR = non‐proactive),  
\textbf{Three-class}  (BA = balanced; PR = proactive; RE = reactive).}
\label{tab:results2}

{\small
\resizebox{\textwidth}{!}{%
\begin{tabular}{@{}lcccccccccc@{}}
\toprule
  & \multicolumn{5}{c}{\textbf{Five‐class}}
  & \multicolumn{2}{c}{\textbf{Binary}}
  & \multicolumn{3}{c}{\textbf{Three‐class}} \\
\cmidrule(lr){2-6} \cmidrule(lr){7-8} \cmidrule(lr){9-11}
& BA & H\_PR & H\_RE & P\_PR & P\_RE & PR & N\_PR & BA & PR & RE \\
\midrule
BT-OC-OP                  & 42.62 & 89.47 & 4.76  & 19.19 & 68.72 & 82.14 & 92.10 & 53.41 & 92.10 & 73.68 \\
BT-OP           & 37.06 & 87.77 & 13.79 & 25.28 & 66.40 & 82.76 & 92.19 & 46.92 & 92.19 & 66.42 \\
BT-BC-OC-OP& 58.24 & 87.84 & 19.05 & 11.82 & 71.75 & 83.48 & 92.90 & 51.67 & 92.90 & 72.09 \\
BT-BC-OP            & 54.74 & 88.89 & 17.39 & 11.00 & 73.68 & 82.97 & 92.58 & 51.76 & 92.58 & 69.57 \\
BC-BT-OC-OP         & 57.06 & 87.73 & 14.81 & 7.07  & 74.89 & 82.59 & 92.23 & 45.00 & 92.32 & 69.96 \\
\bottomrule
\end{tabular}%
}
}
\end{table*}

\begin{figure}[htbp]
    \centering
    \includegraphics[width=0.5\textwidth]{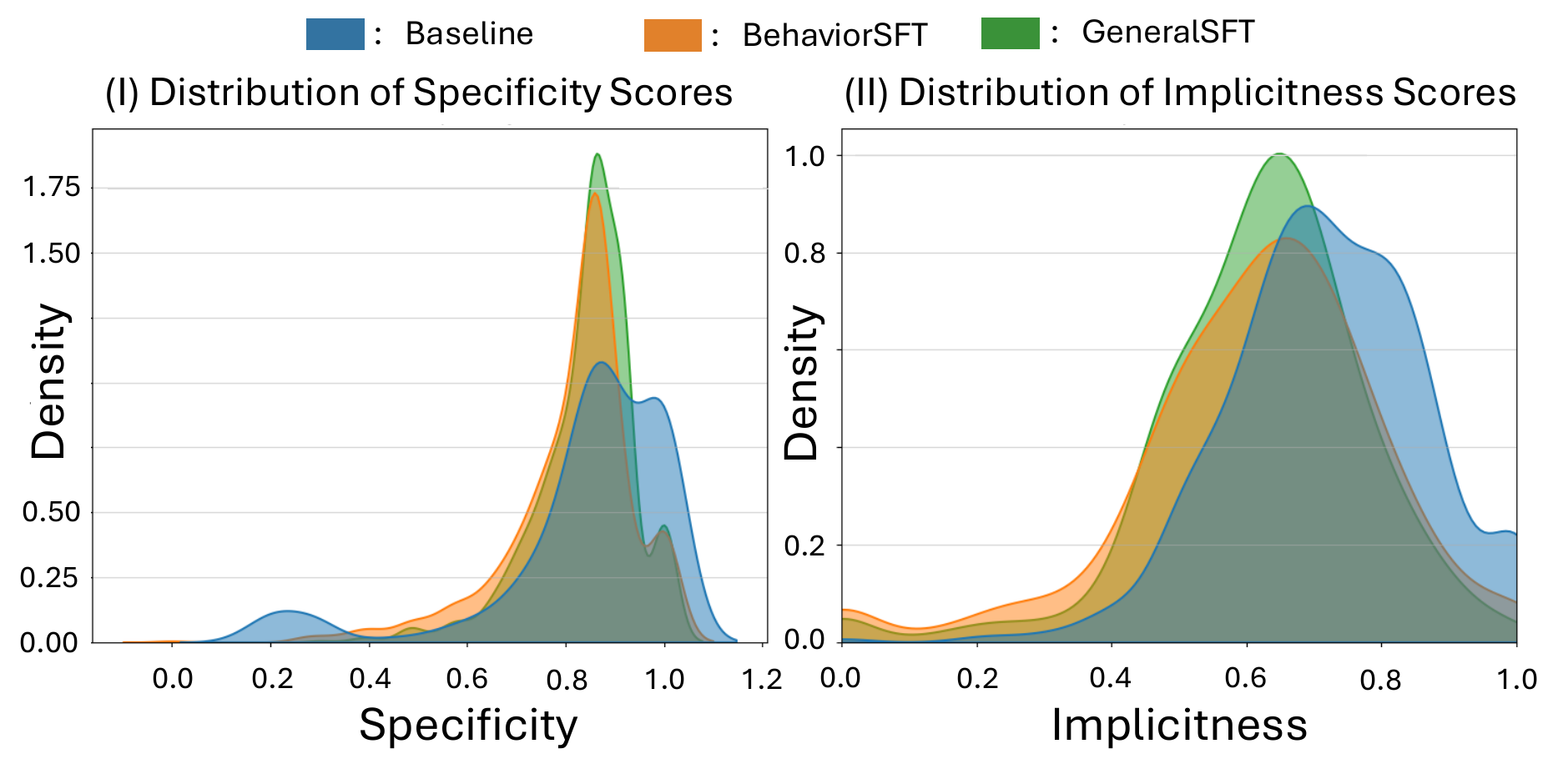} 
    \caption{
        \textbf{Density distributions of (I) Specificity and (II) Implicitness scores for Baseline, BehaviorSFT, and GeneralSFT agent outputs.}
        (I) Specificity: Both fine-tuned models (BehaviorSFT and GeneralSFT) markedly improve output specificity over the Baseline, with distributions concentrated at high scores ($\sim$0.9).
        (II) Implicitness: Distinct implicitness profiles emerge: GeneralSFT is the most explicit (lowest scores, $\sim$0.6-0.7), the Baseline is the most implicit (highest scores, $\sim$0.7-0.9), while BehaviorSFT exhibits a moderate, intermediate level of implicitness ($\sim$0.7-0.8).
    }
    \label{fig:specificity_implicitness_distributions}
\end{figure}

\begin{figure}[htbp]
    \centering
    \includegraphics[width=0.5\textwidth]{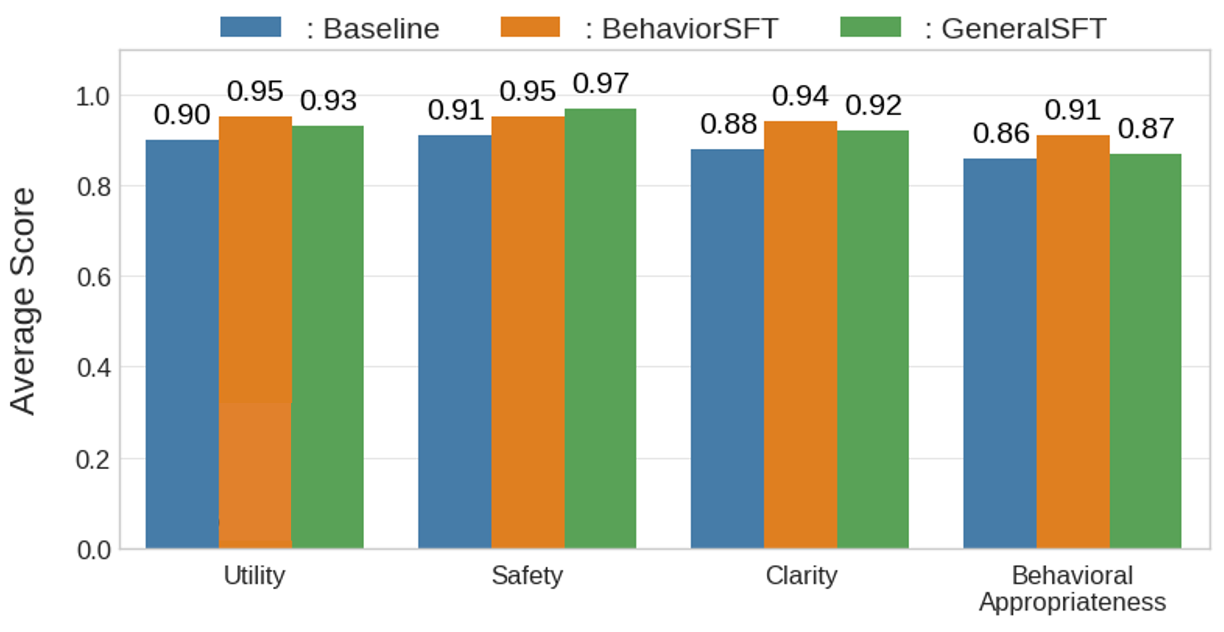}
    \caption{\textbf{G-Eval with \texttt{gpt-4o-mini} as evaluator of \texttt{Qwen-2.5-7B-Ins} responses across four key metrics.} We compare the average scores for the Baseline model, our proposed BehaviorSFT, and GeneralSFT. BehaviorSFT consistently outperforms the Baseline across all metrics and demonstrates competitive or superior performance compared to GeneralSFT.}
    \label{fig:geval_qwen_comparison}
\end{figure}

\section{Experiments and Results}
\label{sec:experiments_results} 

\subsection{Setup}

All experiments use \methodname\ with a fixed \texttt{6\,776/110/977} train–val–test split.  
We fine–tune both backbones; \texttt{Qwen‐2.5‐7B‐Instruct} \cite{qwen2.5} and \texttt{Meta‐Llama‐3.1‐8B‐Instruct} \cite{meta2024llama31}. Details implementation details can be found in Appendix \ref{sec:implementation_details}.



\subsection{Main Results}

\textbf{From Reactive to Proactive capabilities} in clinical LLMs involve processing and responding directly to explicitly provided information. Reactivity encompasses \textit{fact retrieval}, \textit{information summarization}, ordering events via \textit{direct sequencing}, following \textit{simple execution} instructions, and performing \textit{basic reasoning from explicit data}, these tasks test the LLM's ability to understand and manipulate information as presented, without significant inference or applying external knowledge. The Proactive-Reactive Scale of 0.0-0.4 typically reflects these functions.

Conversely, require the LLM to transcend literal interpretation, demonstrating deeper reasoning, anticipation, and critical assessment. Key aspects include \textit{inference and implication} (identifying unstated assumptions or missing information), \textit{anticipation and prediction} (foreseeing next steps or complications), \textit{consistency and conflict detection} (finding discrepancies between data points), \textit{error recognition and correction}, \textit{applying external knowledge} like standards of care, and \textit{synthesis and complex interpretation} from multiple sources. These tasks simulate higher-order clinical thinking. The Proactive-Reactive Scale of 0.6-1.0 aligns with these skills, while 0.4-0.6 represents a balance.

\textbf{Empirical Results Overview.} \quad
Table \ref{tab:main_results1} reports Macro F1 scores across the three task categories. Relative to both the majority-voting \textbf{Ensemble} baseline and standard supervised fine-tuning (\textbf{Gen.~SFT}), \textbf{BehaviorSFT} matches or slightly exceeds performance on the \textit{Reactive} and \textit{Balanced} sets, and yields a clear advantage on the most demanding \textit{Proactive} tasks (Qwen: 96.5\% vs.~95.0\%; Llama: 94.7\% vs.~94.2\%). These gains confirm that the behavior-aligned fine-tuning strategy is particularly effective for higher-order reasoning tasks such as complex inference, error correction, and guideline-based decision making, thereby strengthening the model’s proactive capabilities. Detailed accuracy figures for the three commercial baselines are provided in Appendix \ref{app:baseline_results}.

\textbf{Enhanced User-Centric Qualities with G-Evaluation} \quad
Our evaluation using G-Eval \cite{liu2023g}, a methodology leveraging large models for human-aligned assessment, reveals significant qualitative improvements with BehaviorSFT. As depicted in Figure \ref{fig:geval_qwen_comparison}, BehaviorSFT consistently outperforms the Baseline across all four key metrics: Utility, Safety, Clarity, and Behavioral Appropriateness. Notably, BehaviorSFT achieves the highest scores in Utility (0.95 vs. 0.93 for GeneralSFT and 0.90 for Baseline), Clarity (0.94 vs. 0.92 for GeneralSFT and 0.88 for Baseline), and Behavioral Appropriateness (0.91 vs. 0.87 for GeneralSFT and 0.86 for Baseline). While GeneralSFT scores marginally higher in Safety (0.97 vs. 0.95 for BehaviorSFT), BehaviorSFT still demonstrates a strong safety profile. These results underscore BehaviorSFT's capability to not only perform tasks effectively but also to align more closely with user expectations in terms of usefulness, understandability, and appropriate interaction, suggesting a more refined and user-centric agent behavior.

\textbf{Optimizing Output Specificity while Balancing Implicitness}
Figure \ref{fig:specificity_implicitness_distributions} illustrates the impact of our fine-tuning approaches on the nuanced characteristics of agent responses, specifically their specificity and implicitness. Both fine-tuned models, BehaviorSFT and GeneralSFT, markedly enhance output specificity compared to the Baseline, with distributions concentrating at high specificity scores (around 0.9). This indicates that both methods generate more detailed and precise information. However, a key distinction emerges in their implicitness profiles. GeneralSFT tends towards more explicit communication, reflected in lower implicitness scores (approximately 0.6-0.7). In contrast, the Baseline model is the most implicit (scores around 0.7-0.9). BehaviorSFT carves out an intermediate and potentially more versatile profile, achieving a moderate level of implicitness (scores approximately 0.7-0.8). This suggests that BehaviorSFT can deliver highly specific information without resorting to excessive explicitness, potentially mirroring more natural human communication patterns and aligning with the idea that effective agents must navigate implicit evaluation criteria \cite{wadhwa2025evalagent}. 


\subsection{Ablation on prompting variants for Behavior Pattern Analysis}

Table~\ref{tab:results2} evaluates five prompting recipes obtained by incrementally adding \emph{Behavior Chain-of-Thought} (BC) and \emph{Option reasoning} (OC/OP) on top of the \emph{Behavior Token} (BT) baseline.  
The full recipe \textit{BT--BC--OC--OP} achieves the best or second-best Macro F1 in 11 of the 13 columns (e.g., \textit{Five-class\,BA} 58.2 and \textit{Binary\,PR} 83.5), showing that BC and OC/OP provide complementary gains.  
Dropping OC/OP (\textit{BT--BC--OP}) or BC (\textit{BT--OP}) consistently lowers scores, while reversing the BC placement (\textit{BC--BT--OC--OP}) yields a smaller benefit, indicating that BC is most effective when appended after the BT prompt. Overall, combining both reasoning cues delivers the most robust behaviour classification across all label granularities.





\section{Conclusion}
This paper addresses the critical gap in LLM proactivity for healthcare. Our \textbf{\methodname}, validated by clinicians for plausibility, systematically evaluates this, revealing LLM deficiencies in proactive reasoning despite reactive strengths. We introduced \textbf{BehaviorSFT}, a new fine-tuning method using explicit <reactive> and <proactive> tokens. BehaviorSFT improved performance, achieving up to 97.3\% overall Macro F1 on \methodname and boosting proactive task scores (e.g., Qwen2.5-7B-Ins from \textbf{95.0\% to 96.5\%}). Crucially, in a clinician user study, BehaviorSFT-trained agents received the most favorable rankings (best mean rank \textbf{1.80}). G-Eval results also showed superior Utility (\textbf{0.95}) and Behavioral Appropriateness (\textbf{0.91}). These combined findings demonstrate BehaviorSFT's effectiveness in creating more reliable, clinically nuanced, and clinician-preferred LLM agents for complex healthcare scenarios.

\bibliography{acl_latex}

\appendix

\section{Related Works}

\paragraph{The Evolving Role of AI in Clinical Tasks}
Early AI applications in healthcare predominantly functioned as reactive tools, such as information retrieval systems responding to explicit queries \cite{Yasunaga2022LinkBERT} or basic clinical decision support (CDS) systems triggering alerts based on predefined rules. These systems, while valuable, often lacked contextual understanding and the ability to anticipate clinician needs or potential issues proactively \cite{mccoy2014clinical, sutton2020overview}. More recent advancements, particularly with LLMs, have paved the way for more sophisticated AI assistants. Models like Med-PaLM \cite{singhal2023large} and Med-Alpaca \cite{han2023medalpaca} demonstrated strong domain knowledge, though primarily in a reactive question-answering capacity. The trend is now shifting towards systems with proactive capabilities. For instance, MediQ \cite{li2024mediq} explores proactive information-seeking when context is incomplete, while systems like AIME \cite{tu2024towards} and MDAgents \cite{kim2024mdagents} begin to suggest next steps or anticipate patient needs. 
Our work builds on this trajectory by focusing on systematically training and evaluating the adaptation of reactive and proactive behaviors.

\paragraph{Challenges of Proactive AI in Healthcare} Proactive behaviors in healthcare AI are diverse and critical. One key form is \textit{proactive alerting}, where systems identify and flag critical information, potential errors (e.g., drug interactions, missed standard protocols), or deviations from normal (e.g., critical lab values) \citep{wright2018reduced, fixler2023alert, lee2014medical}. While potentially life-saving, a major challenge is alert fatigue, where excessive or irrelevant alerts lead to high override rates and desensitization among clinicians \citep{gani2025understanding, olakotan2020evaluating, hussain2019medication}. Recent efforts focus on contextualizing alerts to improve relevance and reduce fatigue \citep{poly2020machine, van2021optimizing}. Another crucial area is \textit{proactive information-seeking} under uncertainty. Clinical scenarios often involve incomplete information, and an AI agent should ideally recognize knowledge gaps and ask clarifying questions rather than proceeding with potentially unsafe assumptions \citep{li2024mediq}. \citep{zhang2023clarify} proposed a clarification framework that uses an entropy-based metric to decide when to intervene, improving performance particularly in ambiguous cases. Li et al.\ \citep{li2025two} developed a two-stage dialogue model where the AI actively asks diagnostic questions before refining them, closely emulating physician-like inquiry. Finally, \textit{contextual intervention and suggestion} involve AI volunteering relevant, unprompted information, suggesting next steps, or adapting guidance based on inferred clinical context, user expertise, or workflow stage \citep{widmer2015digital, friend2023wearable, mahajan2025wearable, khalifa2024artificial}. This can manifest as just-in-time proactive guidance \citep{chiou2020harnessing, gebreab2024llm}. The core challenge, which our work directly addresses, is adapting \textit{when} and \textit{how} to intervene to be helpful without being disruptive or unsafe \citep{fauscette2024agentic}.

\paragraph{Controllable Generation for Healthcare LLMs}
Controlling the behavior of LLMs beyond simple task completion is an active research area. Techniques range from inserting learnable control signals like prefix-tuning or using special tokens \citep{goyal2023think, dathathri2019plug} to preference-based fine-tuning (e.g., RLHF) to encourage specific interaction styles. Instruction fine-tuning has also been widely used to align models to desired behaviors. \citep{chen2023controllable} showed that large LMs can adopt initiative-taking or supportive dialogue strategies through prompt design alone, without additional model tuning. Several benchmarks exist for evaluating LLMs in medicine, such as MedQA \citep{jin2021disease}, PubMedQA \citep{jin2019pubmedqa}, MedMCQA \citep{pal2022medmcqa}, and more recent ones like MedAgentBench \citep{jiang2025medagentbench} or ClinicBench \citep{chen2024clinicalbench}. These primarily focus on knowledge accuracy, reasoning over medical facts, or agentic task completion. While some, like MediQ \citep{li2024mediq}, touch upon aspects of proactivity (information-seeking), there is a lack of systematic frameworks to evaluate and train LLMs specifically on their ability to dynamically adapt their behavior along the full reactive–proactive spectrum in diverse clinical contexts. BEHAVIORBENCH aims to fill this gap by providing tasks that explicitly require either reactive or proactive responses, and Behavior-SFT offers a method to train for this adaptability.

\section{Limitations and Future Works} 

\vspace{-4pt}

\paragraph{Data \& Task Scope.}
\textsc{BehaviorBench} aggregates 6,876 English clinical vignettes (142K task instances) from \emph{NEJM}. This corpus reflects an internal‑medicine bias and omits modalities such as radiology reads, nursing shift notes, tele‑health transcripts, and non‑English documentation. The future tasks include expanding the benchmark to multilingual EHR snippets and image‑grounded prompts, and we are adding tasks for dermatology, psychiatry, and longitudinal trend summarisation to test whether proactive cues generalise beyond text‑only, single‑visit encounters.

\paragraph{Behaviour Modelling.}
Our \textsc{BehaviorSFT} controller currently toggles generation with a binary \texttt{<reactive>} / \texttt{<proactive>} token.  
Although effective for coarse behaviour shifts, this switch cannot express nuances such as anticipatory clarification versus high‑urgency escalation, and it occasionally over‑fires, creating alert fatigue.  
We are experimenting with a hierarchical token inventory (e.g.\ \texttt{<clarify\_info>}, \texttt{<flag\_safety>}, \texttt{<escalate\_critical>}) learnt from multi‑label supervision, and with behaviour‑weighted RLHF that continuously trades helpfulness against cognitive load.

\paragraph{Evaluation \& Deployment Readiness.}
The clinician study in Appendix \ref{sec:user_study} involves three medical doctors number of cases sufficient for validation but under‑powered for robust error stratification or workflow integration. Future work should recruit multi‑institution cohorts (20+ clinicians, 1,000+ cases) and embeds the agent inside a simulated EHR sandbox to observe interrupt patterns, hand‑off continuity, and long‑horizon reasoning across multi‑day episodes.

\section{Ethical Implications}
\vspace{-4pt}

\paragraph{Safety \& Accountability.}
Proactive agents can prevent omission errors, yet incorrect or over‑confident interventions may induce \emph{commission} errors that are harder to detect.  
We therefore plan to release model checkpoints after careful reviews.  
Post‑deployment, we advocate continuous monitoring with an audit trail that logs every proactive trigger and its downstream clinical action for root‑cause analysis.

\paragraph{Fairness \& Bias Mitigation.}
Because benchmark data are skewed toward North‑American populations, behaviour triggers may under‑fire on minority phenotypes or over‑fire on stigmatised conditions, reinforcing disparities.  
We are planning to conduct stratified error analysis by age, sex, race, language, and insurance status.  
Future releases will contain group‑specific performance cards and debiasing adapters that minimise disparate false‑negative / false‑positive rates while preserving recall on the majority group.

\paragraph{Data Privacy \& Responsible Release.}
All medical cases are available for those institutions who purchased NEJM license; nonetheless, fine‑tuned models might memorize private strings when trained on institutional EHRs. We will publish an \textbf{Ethical Usage Card} outlining intended tasks, known failure modes, monitoring hooks, and sunset clauses for model retirement, and we encourage downstream users to adopt the same safeguards.

\section{Dataset Statistics}
\label{sec:dataset_statistics}
The final \methodname dataset consists of 6,876 real-world clinical case scenarios from which we derived a total of 142,496 tasks distributed across the 13 distinct task categories described in Section~\ref{sec:behaviorbench}. 

\subsection{Simulated Conversations}
The simulated conversations in the \methodname dataset are derived from real-world clinical case reports published in the New England Journal of Medicine (NEJM). Each conversation reconstructs the clinical reasoning process among healthcare professionals, encompassing diagnostic deliberation, treatment planning, and communication with patients and caregivers.

\begin{figure*}[h!]
    \centering
    \includegraphics[width=0.7\textwidth]{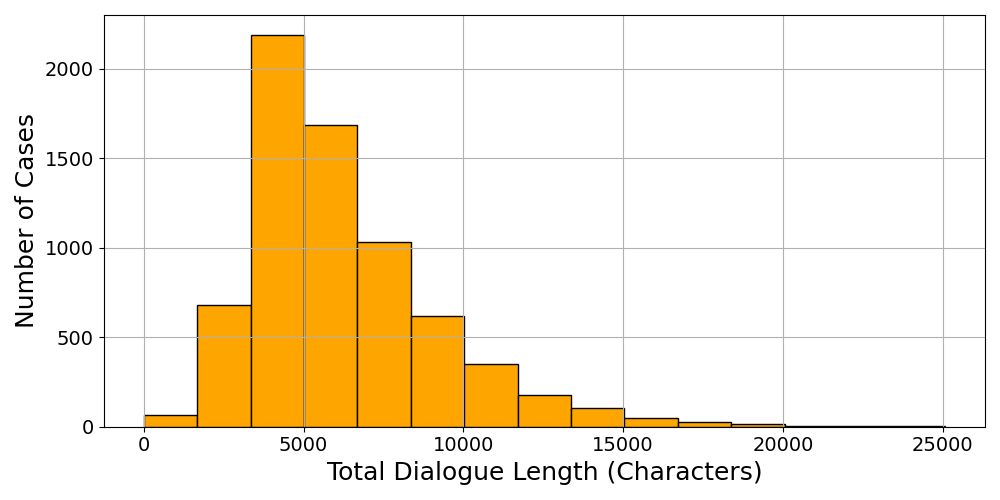}
    \caption{\textbf{Distribution of total dialogue length (in characters) per conversation.} This metric captures the overall verbosity of clinical discussions. Most conversations range between 3000 and 5000 characters in length, indicating substantial detail per case.}
    \label{fig:dialogue_length_dist} 
\end{figure*}

\begin{figure*}[h!]
    \centering
    \includegraphics[width=0.7\textwidth]{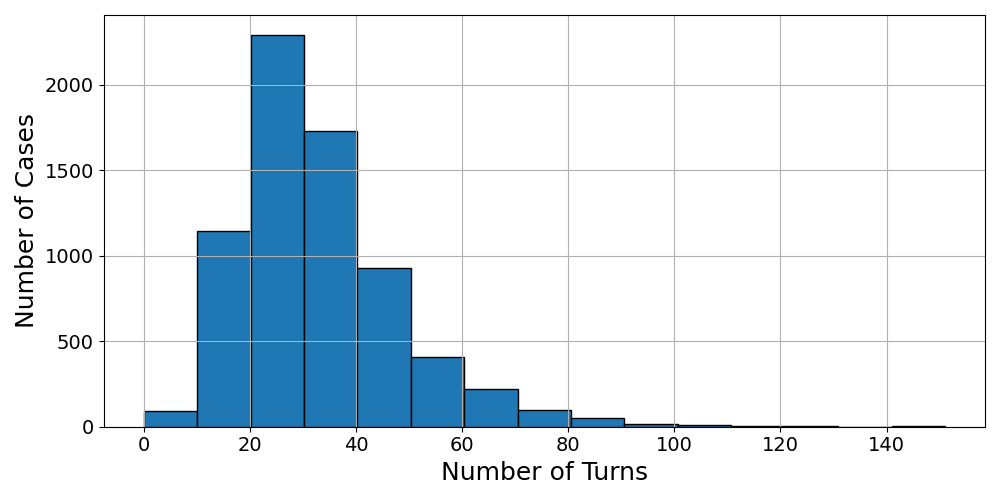}
     \caption{\textbf{Distribution of the number of dialogue turns per conversation.} Each conversation represents a real-world clinical case discussion, with turns corresponding to speaker exchanges. The majority of cases fall between 15 and 30 turns.}
    \label{fig:num_turn_dist} 
\end{figure*}

Table~\ref{tab:conversation_summary_stats} and Figure ~\ref{fig:dialogue_length_dist} and \ref{fig:num_turn_dist} provide descriptive statistics of the conversation data, illustrating the natural variability and complexity of the simulated dialogues. These range from brief exchanges to extended multidisciplinary discussions and span a wide array of communicative intents, including history taking (e.g., eliciting chief complaint, symptom duration, and past medical history), physical examination interpretation, diagnostic reasoning, and family updates. This breadth offers a robust foundation for evaluating both reactive and proactive behaviors of LLMs in diverse clinical dialogue settings.

\begin{table}[ht]
\centering
\caption{\textbf{Summary Statistics of Simulated Clinical Conversations}. This table reports average structural properties of the conversations in the dataset, including the number of dialogue turns, total dialogue length in characters, and number of unique participants per case.}
\label{tab:conversation_summary_stats}
\begin{tabular}{lc}
\toprule
\textbf{Metric} & \textbf{Value} \\
\midrule
Avg. \# of turns per conversation & 33.3 \\
Avg. len of dialogue per conversation & 6194.3 \\
Avg. \# of participants per case & 8.7 \\
\bottomrule
\end{tabular}
\end{table}

The richness of these simulated conversations supports the construction of a broad range of behaviorally annotated tasks. These tasks underpin our evaluation framework, which is designed to assess not only reactive capabilities, such as information retrieval, but also proactive competencies such as anticipatory reasoning and clinical foresight.

\subsection{Tasks}
\begin{figure*}[h!] 
    \centering
    \includegraphics[width=\textwidth]{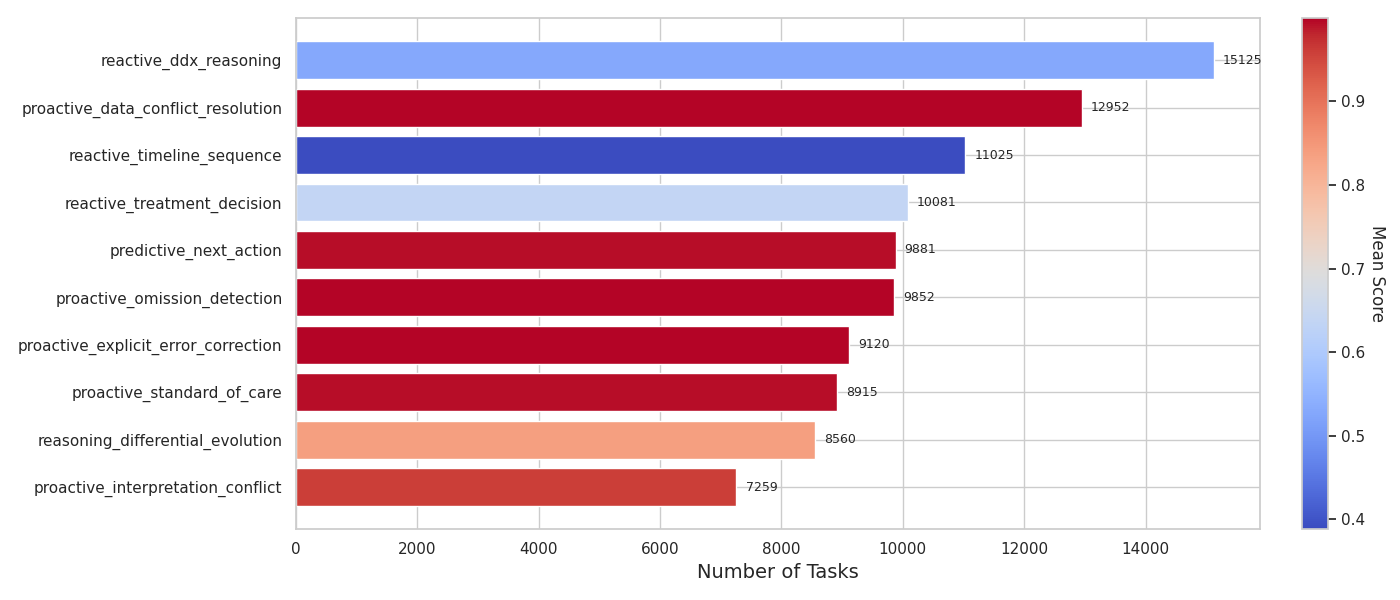}
   \caption{\textbf{Distribution of instances across specific task types in \methodname.} Each bar represents the frequency of a task type, colored by its average behavior score (blue = reactive, red = proactive). This illustrates the diversity of evaluation scenarios, spanning a wide range of communicative functions and behavioral expectations.}

    \label{fig:task_dist} 
\end{figure*}

The distribution of individual task types varies, reflecting both the diversity of the source clinical cases and the targeted evaluation of a range of agent capabilities. Figure~\ref{fig:task_dist} presents detailed counts for the ten most prevalent task types.

\begin{figure*}[h!]
    \centering
    \includegraphics[width=0.7\textwidth]{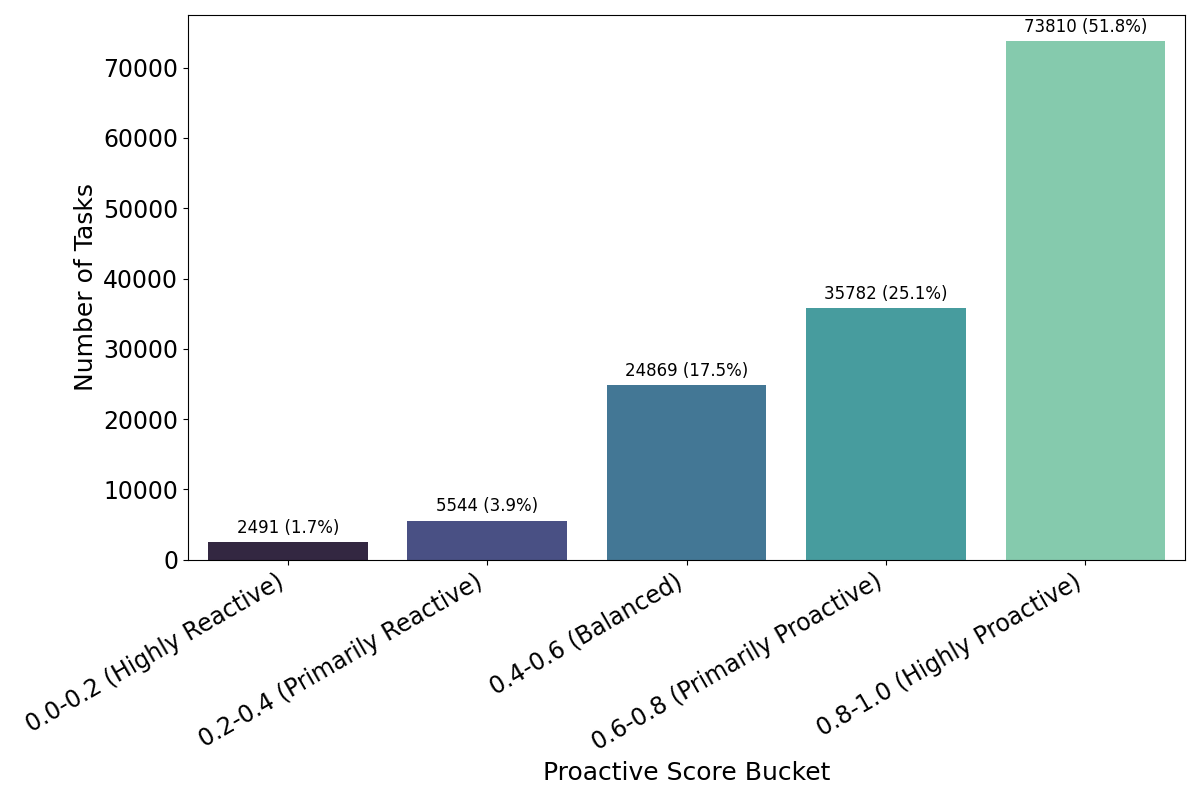}
\caption{\textbf{Distribution of tasks across discrete behavior categories in \methodname.} Tasks are grouped into five categories, ranging from `highly reactive' to `highly proactive' to support structured evaluation of agent behavior along the autonomy spectrum.}

    \label{fig:proactive_category_dist} 
\end{figure*}

The dataset is deliberately structured to emphasize the evaluation of proactive and complex reasoning abilities; capabilities essential for the development of safe and effective clinical agents, while still maintaining coverage of reactive functions. This emphasis is evident in the distribution across broader behavioral categories (Appendix Figure~\ref{fig:proactive_category_dist}): the largest group comprises \textit{highly proactive} tasks (73,810 instances), followed by \textit{primarily proactive} tasks (35,782 instances). \textit{Primarily reactive} (5,544 instances) and \textit{highly reactive} (2,491 instances) tasks ensure comprehensive coverage of reactive tasks. Additionally, \textit{balanced} tasks (24,869 instances) ensure that the full spectrum is represented.

\begin{figure*}[h!]
    \centering
    \includegraphics[width=0.7\textwidth]{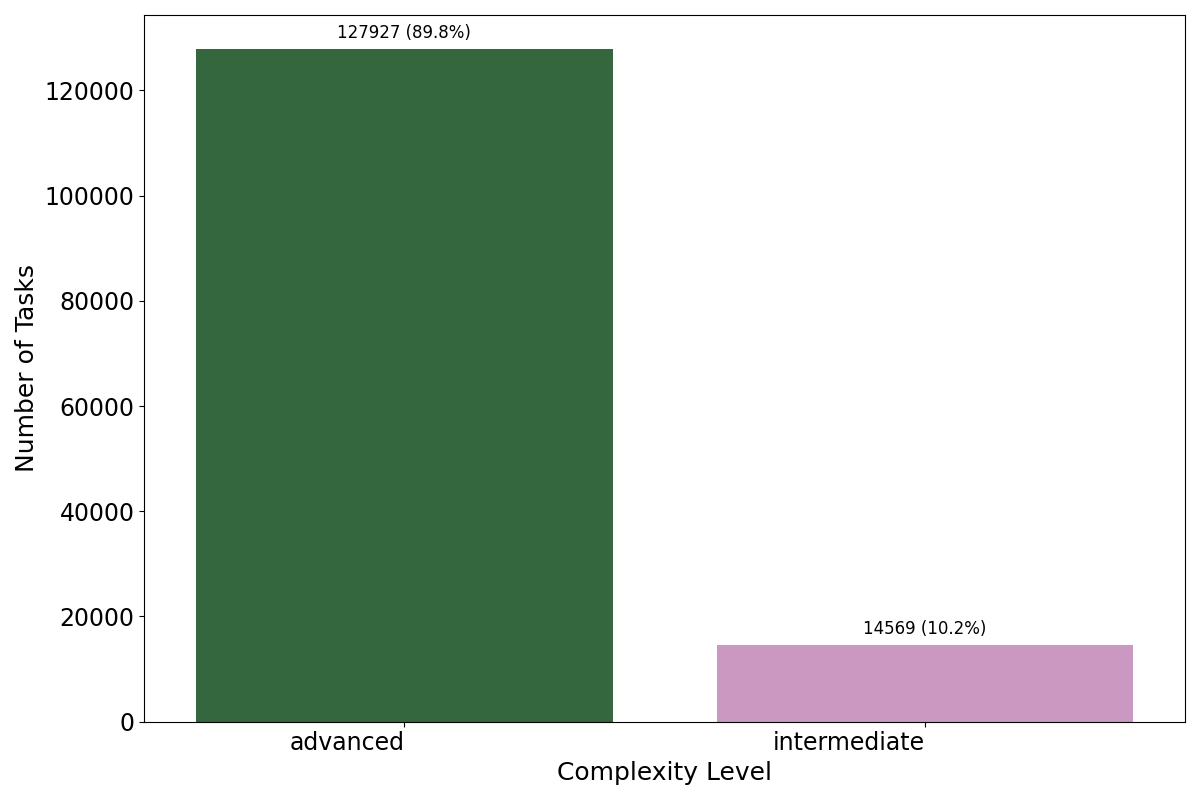}
\caption{\textbf{Distribution of instances by task complexity level in \methodname.} Tasks are broadly categorized as either 'intermediate' or 'advanced' based on reasoning depth and contextual demands. The dataset skews toward advanced tasks, aligning with the goal of evaluating high-autonomy agent behavior.}

    \label{fig:complexity_dist} 
\end{figure*}

We also categorize tasks by complexity, broadly distinguishing between `intermediate' tasks (often corresponding to simpler reactive functions) and `advanced' tasks (typically involving proactive or complex balanced reasoning). The dataset heavily features `advanced' tasks (127,927 instances) compared to `intermediate' tasks (14,569 instances), as shown in Figure \ref{fig:complexity_dist}, where the advanced tasks feature a higher proactive score of above 0.8 compared to intermediate tasks with an average of 0.4 proactive score (Figure \ref{fig:avg_tokens_complexity} in Appendix).

Furthermore, a continuous behavior score (ranging from 0.0 for fully reactive to 1.0 for fully proactive, defined in Section \ref{sec:experiments_results}) was assigned during annotation. The distribution of these scores (Figure \ref{fig:proactive_score_dist} in Appendix) shows a concentration towards higher proactivity (0.6-1.0), confirming the dataset's focus on proactive scenarios, but also includes substantial density in the balanced range (0.4-0.6) and coverage of reactive cases (0.0-0.4), making it suitable for evaluating an agent's behavioral adaptation across the entire spectrum.

\section{The Landscape of Healthcare AI}

\begin{figure*}[t!]
    \centering\includegraphics[width=1.0\textwidth]{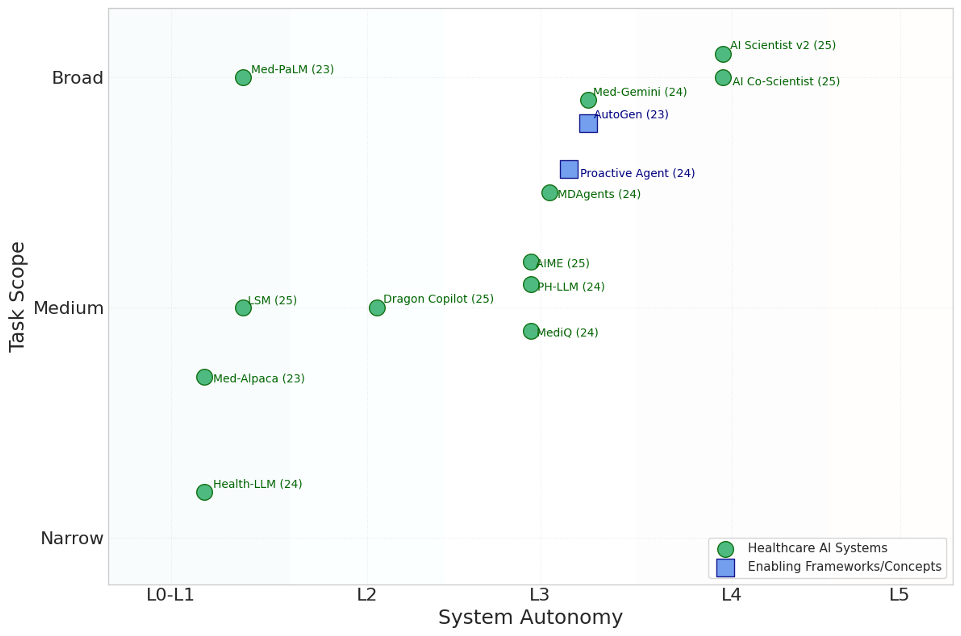}
    \caption{\textbf{The Landscape of Healthcare AI Systems and Enabling Frameworks.} Systems are positioned based on their primary Task Scope (Narrow, Medium, or Broad) and their demonstrated level of System Autonomy. The autonomy levels are derived from the Six-Level Taxonomy for Healthcare AI Agent Autonomy (detailed in Table~\ref{tab:autonomy_taxonomy_detailed}), ranging from L0-L1 (Assistance \& Reactive Info) through L3 (Conditional Automation/Contextual Proactivity) to L4-L5 (High/Full Automation). Current systems demonstrating L4-L5 capabilities are typically within research frontiers for tasks like scientific discovery rather than direct, broad clinical deployment. Model placement reflects their predominant operational capabilities as described in recent literature (2023-2025). The progression towards higher autonomy, particularly the transition from L2 (Reactive Support) to L3 (Contextual Proactivity), necessitates significant advancements in behavioral adaptation to ensure safe and effective operation in nuanced healthcare contexts. Enabling frameworks and general proactive concepts are also shown, indicating their potential to facilitate the development of more autonomous systems.}
    \label{fig:behavioral_adaptation} 
\end{figure*}

The capabilities of Artificial Intelligence (AI) systems in healthcare are rapidly advancing, moving beyond simple information retrieval towards more autonomous and complex task handling. Figure~\ref{fig:behavioral_adaptation} provides a visual representation of this evolving landscape, positioning various contemporary Healthcare AI Systems and Enabling Frameworks/Concepts based on two key dimensions: their operational Task Scope and their level of System Autonomy.

The System Autonomy axis is rigorously grounded in the Six-Level Taxonomy for Healthcare AI Agent Autonomy (detailed in Table~\ref{tab:autonomy_taxonomy_detailed} in the Appendix). This taxonomy delineates capabilities from Level 0-1 (No Automation/Clinician Assistance), where AI provides reactive information or simple alerts, through Level 2 (Partial Automation/Reactive Support), where AI executes specific clinician-commanded tasks.

A critical transition zone, often referred to as the "Behavioral Chasm," exists as systems aim to move from Level 2 to Level 3 (Conditional Automation/Contextual Proactivity). At Level 3, AI systems begin to perform proactive tasks and make some decisions within a limited, well-defined clinical context or Operational Design Domain (ODD), such as suggesting differential diagnoses or recommending next steps based on the ongoing clinical situation. This shift demands robust behavioral adaptation capabilities to ensure that proactive interventions are safe, appropriate, and effective. Our work on BehaviorSFT and the BehaviorBench evaluation framework is specifically aimed at addressing the challenges of training and assessing these crucial Level 3 behaviors, which are vital for the development of reliable AI co-pilots and assistants. As illustrated in Figure~\ref{fig:behavioral_adaptation}, many contemporary applied systems such as MediQ \citep{li2024mediq}, AIME \citep{tu2024towards}, and Med-Gemini \citep{saab2024capabilities} are operating at or pushing the boundaries of Level 3 capabilities.

The higher autonomy levels, L4 (High Automation/Proactive Decision Support) and L5 (Full Automation/Autonomous Operation), represent the current research frontier for AI in healthcare. Systems like AI Co-Scientist \citep{gottweis2025towards} and AI Scientist v2 \citep{yamada2025ai}, while focused on scientific discovery, demonstrate capabilities that conceptually align with L4 by making significant decisions and taking proactive actions within their research ODDs with minimal human oversight for extended periods. Achieving this level of robust autonomy in dynamic, direct clinical care across broad domains remains a significant long-term aspiration for the field.

Enabling frameworks such as AutoGen \citep{wu2023autogen} and general concepts like the Proactive Agent \citep{lu2024proactive} are instrumental in this progression. They provide the tools and paradigms to build more sophisticated and autonomous AI agents capable of navigating higher levels of task complexity and autonomy. The continued development in this field underscores the critical importance of ensuring that as AI systems become more autonomous, their behaviors are rigorously evaluated and remain aligned, safe, and beneficial within the complex and high-stakes domain of healthcare.

\section{Baseline Performance}
\label{app:baseline_results}

Tables~\ref{tab:main_results2}, \ref{tab:main_results3}, and \ref{tab:main_results4} compare \texttt{o1},
\texttt{Gemini-2.5 Pro}, and \texttt{DeepSeek-R1} under three prompting regimes---Zero-Shot (ZS), Few-Shot with three examples (FS), and ZS augmented by explicit reactive/proactive instructions. All models score near-ceiling on the \emph{Reactive} and \emph{Balanced} subsets, but diverge on the harder \emph{Proactive} tasks, where \texttt{DeepSeek-R1} attains the highest average accuracy (95\%), edging out \texttt{Gemini} and \texttt{o1} (both~\(\approx 93\%\)). Across models, FS generally yields the most consistent gains; especially on items such as \emph{predictive next action}, while explicit instructions benefit \texttt{DeepSeek} yet can slightly reduce performance for \texttt{Gemini} and \texttt{o1}. These results underscore that, although lower-level clinical reasoning is largely saturated, proactive reasoning remains the principal differentiator among state-of-the-art LLMs.

\section{Prompt Template}
\label{sec:prompt_template}
\tikzset{
    mybox/.style={draw=black, very thick, rectangle, rounded corners, inner sep=10pt, inner ysep=13pt},
    fancytitle/.style={fill=black, text=white, rounded corners, inner xsep=7pt, inner ysep=3.5pt} 
}

\vspace{-2em}

\begin{tcolorbox}[
    enhanced,
    breakable, 
    pad at break=2mm,
    colback=white, 
    colframe=black,  
    arc=3mm,         
    boxrule=0.8pt,   
    title=BehaviorSFT Prompt, 
    fonttitle=\bfseries\normalsize, 
    colbacktitle=black, 
    coltitle=white,     
    attach boxed title to top left={
        yshift=-\tcboxedtitleheight/2, 
        xshift=10mm 
    },
    boxed title style={
        arc=2mm, 
        colframe=black, 
        boxrule=0.5pt,
    },
    top=12pt, 
    bottom=8pt,
    left=8pt,
    right=8pt,
]
\footnotesize
\setstretch{1.2}

\vspace{-0.8em}
You are a helpful medical assistant.

\textbf{Medical Information:}\\
The patient's history of present illness includes treatment with salve, Alpine lamp, intravenous and intramuscular injections, and Fowler's solution.

\vspace{0.5em}
\textbf{Question:} \\
Based on the information in the case summary, how did the patient's treatment for his skin condition evolve from the initial presentation of `eczema' to the administration of Fowler's solution (arsenic)?

\vspace{0.5em}

\textbf{Options:}\\
A: "Initially treated with topical steroids... \\
B: "Initially treated with herbal ... \\
....

\vspace{0.5em}

\textbf{Instruction:}\\
According to the previous information, give me the behavior first (highly\_reactive, primarily\_reactive, balanced, highly\_proactive, primarily\_proactive), then the Rationale and answer in <answer></answer>, later is the detailed option.

\end{tcolorbox}

\begin{figure*}[htbp]
    \centering
    \includegraphics[width=0.8\textwidth]{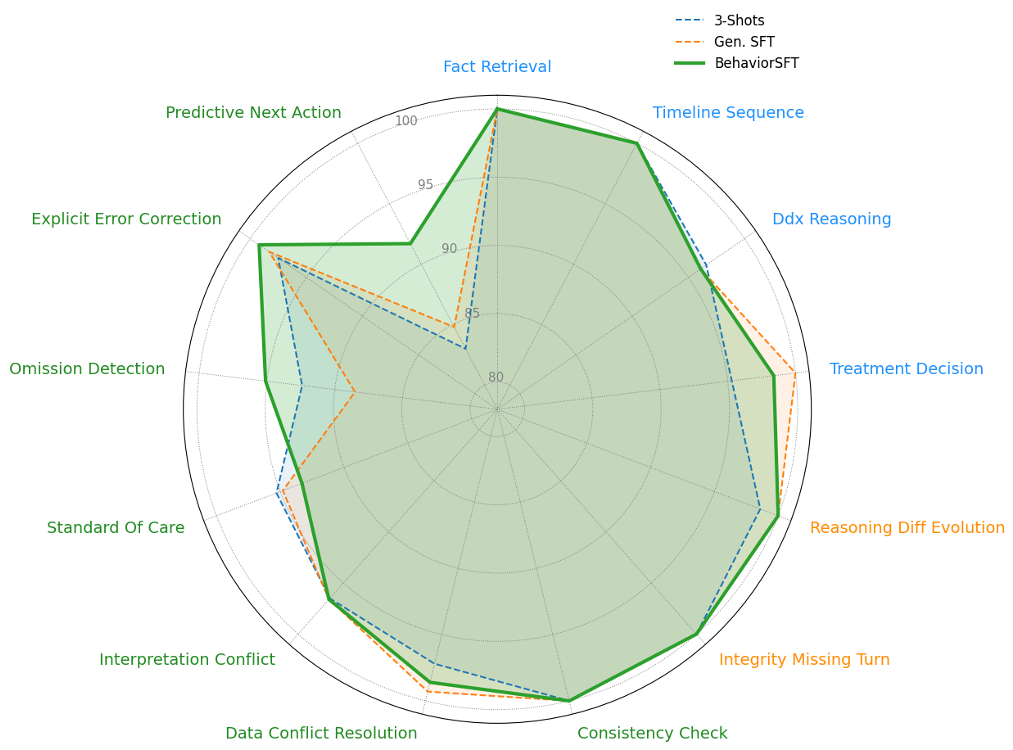}
    \caption{\textbf{Performance comparison on \methodname for Few-Shot (k=3); Gen. SFT, and our proposed BehaviorSFT.} Tasks are colored based on task category: \textcolor{blue}{Reactive}, \textcolor{orange}{Balanced}, and \textcolor{green}{Proactive}. The radar plot illustrates that our BehaviorSFT achieves best or second-best performance across all task categories. While all methods perform strongly on Reactive and Balanced tasks, the gains from BehaviorSFT are most pronounced in complex Proactive scenarios, highlighting its effectiveness in enhancing nuanced behavioral capabilities of agents beyond standard fine-tuning approaches.}
    \label{fig:radar_behavior_sft}
\end{figure*}

\begin{table*}[t!]
\centering
\small
\caption{\textbf{Performance Evaluation on \methodname.}  
Accuracy (\%) across task categories. Best result per task in \textbf{bold}.  
Baseline LLM is \texttt{o1}.  
`ZS' = Zero‑Shot, `FS (k=3)' = Few‑Shot (3 examples),  
`Explicit Instr.' = ZS with explicit reactive/proactive instruction.}
\label{tab:main_results2}
\begin{tabular}{@{}lcccc@{}}
\toprule
\textbf{Category} & \textbf{Task} & \multicolumn{3}{c}{\textbf{Baseline}} \\
\cmidrule(lr){3-5}
 &  & ZS & FS (k=3) & ZS + Explicit Instr. \\
\midrule
\multirow{4}{*}{\rotatebox[origin=c]{90}{\textbf{Reactive}}}
& fact\_retrieval          & \textbf{100.00} & \textbf{100.00} & \textbf{100.00} \\
& timeline\_sequence       & \textbf{100.00} & \textbf{100.00} & \textbf{100.00} \\
& ddx\_reasoning           & \textbf{93.92}  & 91.96 & 91.92 \\
& treatment\_decision      & 91.88 & \textbf{93.78} & 91.88 \\
\cmidrule(lr){2-5}
& \textit{Average}         & \textbf{96.45} & 96.43 & 95.95 \\
\midrule
\multirow{2}{*}{\rotatebox[origin=c]{90}{\textbf{Balanced}}}
& reasoning\_diff\_evolution & 98.05 & \textbf{100.00} & \textbf{100.00} \\
& integrity\_missing\_turn   & \textbf{100.00} & 98.46 & \textbf{100.00} \\
\cmidrule(lr){2-5}
& \textit{Average}           & 99.03 & 99.23 & \textbf{100.00} \\
\midrule
\multirow{7}{*}{\rotatebox[origin=c]{90}{\textbf{Proactive}}}
& consistency\_check         & 95.23 & \textbf{95.24} & 90.12 \\
& data\_conflict\_resolution & \textbf{96.52} & 96.44 & 95.11 \\
& interpretation\_conflict   & \textbf{98.48} & 98.30 & 98.29 \\
& standard\_of\_care         & 91.47 & 91.79 & \textbf{94.87} \\
& omission\_detection        & 81.87 & \textbf{82.00} & 81.61 \\
& explicit\_error\_correction& 96.30 & \textbf{98.12} & 95.54 \\
& predictive\_next\_action   & 78.03 & \textbf{82.88} & 78.30 \\
\cmidrule(lr){2-5}
& \textit{Average}           & \textbf{93.31} & 92.11 & 90.55 \\
\midrule
\multicolumn{2}{@{}l}{\textbf{Average}} & 93.86 & \textbf{94.25} & 93.55 \\
\bottomrule
\end{tabular}%
\end{table*}

\begin{table*}[t!]
\centering
\small
\caption{\textbf{Performance Evaluation on \methodname.} We report Accuracy (\%) across different task categories. Best result per task is highlighted in \textbf{bold}. Baseline LLM used is \texttt{Gemini‑2.5 Pro}. `ZS' = Zero‑Shot, `FS (k=3)' = Few‑Shot (3 examples), `CoT' = Chain‑of‑Thought, `Explicit Instr.' = ZS with explicit reactive/proactive instruction.}
\label{tab:main_results3}
\begin{tabular}{@{}lcccc@{}}
\toprule
\textbf{Category} & \textbf{Task} & \multicolumn{3}{c}{\textbf{Baseline}} \\
\cmidrule(lr){3-5}
& & ZS & FS (k=3) & ZS + Explicit Instr. \\
\midrule
\multirow{4}{*}{\rotatebox[origin=c]{90}{\textbf{Reactive}}}
& fact\_retrieval          & \textbf{100.00} & \textbf{100.00} & \textbf{100.00} \\
& timeline\_sequence       & \textbf{99.10}  & 78.65 & \textbf{99.10} \\
& ddx\_reasoning           & \textbf{95.33}  & 93.99 & 94.56 \\
& treatment\_decision      & \textbf{94.77}  & 93.88 & 94.29 \\
\cmidrule(lr){2-5}
& \textit{Average}         & \textbf{97.30}  & 91.63 & 96.99 \\
\midrule
\multirow{2}{*}{\rotatebox[origin=c]{90}{\textbf{Balanced}}}
& reasoning\_diff\_evolution & \textbf{98.59} & 82.33 & 97.26 \\
& integrity\_missing\_turn   & \textbf{98.46} & 98.05 & 96.56 \\
\cmidrule(lr){2-5}
& \textit{Average}           & \textbf{98.53} & 90.19 & 96.91 \\
\midrule
\multirow{7}{*}{\rotatebox[origin=c]{90}{\textbf{Proactive}}}
& consistency\_check          & 94.29 & \textbf{96.34} & 94.29 \\
& data\_conflict\_resolution  & 97.18 & 97.24 & \textbf{98.53} \\
& interpretation\_conflict    & \textbf{96.70} & 95.11 & 94.95 \\
& standard\_of\_care          & 95.32 & \textbf{96.80} & 92.11 \\
& omission\_detection         & 81.57 & \textbf{90.10} & 79.12 \\
& explicit\_error\_correction & \textbf{96.34} & 94.23 & 95.55 \\
& predictive\_next\_action    & 77.88 & \textbf{81.55} & 73.25 \\
\cmidrule(lr){2-5}
& \textit{Average}           & 91.33 & \textbf{93.05} & 89.69 \\
\midrule
\multicolumn{2}{@{}l}{\textbf{Average}} & \textbf{94.27} & 92.17 & 93.04 \\
\bottomrule
\end{tabular}%
\end{table*}

\begin{table*}[t!]
\centering
\small
\caption{\textbf{Performance Evaluation on \methodname.} We report Accuracy (\%) across different task categories. Best result per task is highlighted in \textbf{bold}. Baseline LLM used is \texttt{DeepSeek‑R1}. `ZS' = Zero‑Shot, `FS (k=3)' = Few‑Shot (3 examples), `CoT' = Chain‑of‑Thought, `Explicit Instr.' = ZS with explicit reactive/proactive instruction.}
\label{tab:main_results4}
\begin{tabular}{@{}lcccc@{}}
\toprule
\textbf{Category} & \textbf{Task} & \multicolumn{3}{c}{\textbf{Baseline}} \\
\cmidrule(lr){3-5}
 &  & ZS & FS (k=3) & ZS + Explicit Instr. \\
\midrule
\multirow{4}{*}{\rotatebox[origin=c]{90}{\textbf{Reactive}}}
 & fact\_retrieval            & \textbf{100.00} & \textbf{100.00} & \textbf{100.00} \\
 & timeline\_sequence         & \textbf{100.00} & \textbf{100.00} & \textbf{100.00} \\
 & ddx\_reasoning             & 93.16 & 91.16 & \textbf{94.25} \\
 & treatment\_decision        & 94.22 & \textbf{95.70} & 94.77 \\
\cmidrule(lr){2-5}
 & \textit{Average}           & 96.84 & 96.71 & \textbf{97.26} \\
\midrule
\multirow{2}{*}{\rotatebox[origin=c]{90}{\textbf{Balanced}}}
 & reasoning\_differential\_evolution & \textbf{98.59} & \textbf{98.59} & \textbf{98.59} \\
 & integrity\_missing\_turn\_inference & \textbf{100.00} & \textbf{100.00} & \textbf{100.00} \\
\cmidrule(lr){2-5}
 & \textit{Average}           & \textbf{99.29} & \textbf{99.29} & \textbf{99.29} \\
\midrule
\multirow{7}{*}{\rotatebox[origin=c]{90}{\textbf{Proactive}}}
 & consistency\_check         & 94.29 & 94.29 & \textbf{100.00} \\
 & data\_conflict\_resolution & 97.18 & 95.68 & \textbf{97.88} \\
 & interpretation\_conflict   & \textbf{100.00} & 96.53 & 98.22 \\
 & standard\_of\_care         & 93.52 & \textbf{95.32} & 94.67 \\
 & omission\_detection        & \textbf{93.78} & 90.75 & 93.57 \\
 & explicit\_error\_correction& 97.50 & 97.52 & \textbf{98.26} \\
 & predictive\_next\_action   & 78.54 & 80.86 & \textbf{82.69} \\
\cmidrule(lr){2-5}
 & \textit{Average}           & 93.54 & 92.99 & \textbf{95.04} \\
\midrule
\multicolumn{2}{@{}l}{\textbf{Average}} & 95.49 & 94.96 & \textbf{96.10} \\
\bottomrule
\end{tabular}%
\end{table*}

\newpage

\begin{figure*}[h!]
    \centering
    \includegraphics[width=0.7\textwidth]{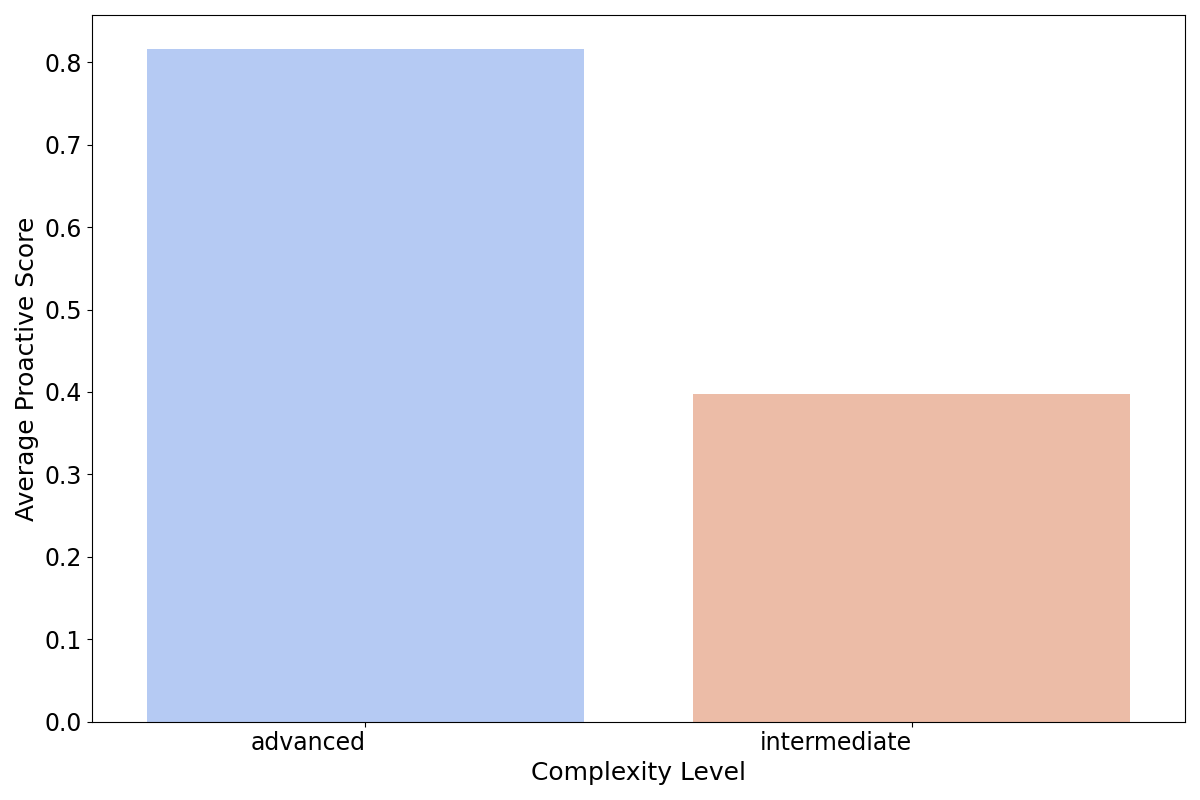}
\caption{\textbf{Average proactive score by task complexity level in \methodname.} Tasks labeled as ‘advanced’ exhibit a significantly higher average proactive score (above 0.8) compared to ‘intermediate’ tasks (around 0.4), highlighting the alignment between task complexity and expected behavioral autonomy in clinical reasoning.}
    \label{fig:avg_tokens_complexity} 
\end{figure*}

\begin{figure*}[h!]
    \centering
    \includegraphics[width=0.7\textwidth]{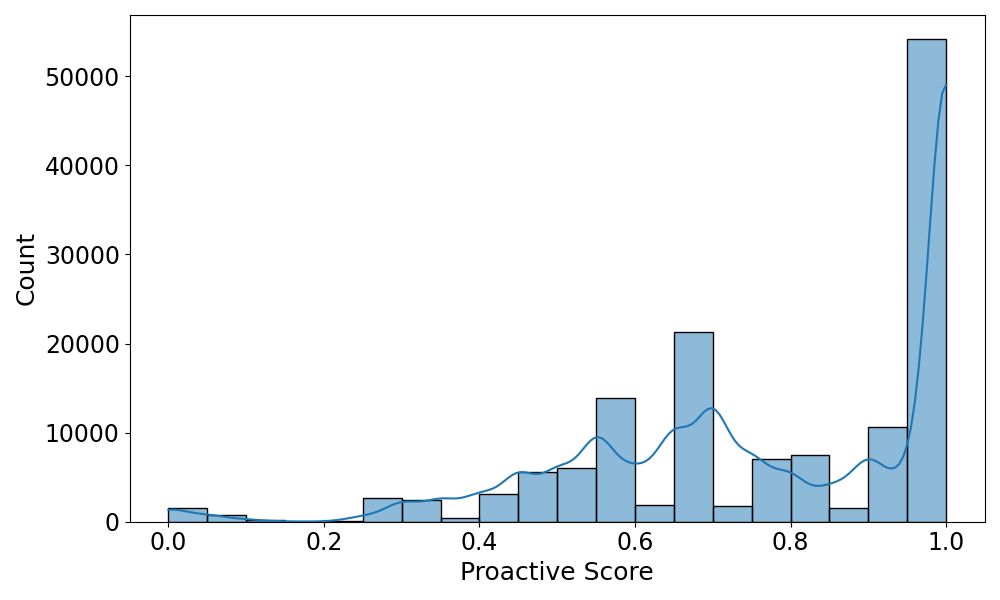}
\caption{\textbf{Distribution of continuous behavior scores across all tasks in \methodname.} The behavior score ranges from 0.0 (fully reactive) to 1.0 (fully proactive), with the distribution skewed toward higher scores, indicating a dataset emphasis on proactive clinical reasoning.}

    \label{fig:proactive_score_dist} 
\end{figure*}

\section{Implementation Details}
\label{sec:implementation_details}

Our BehaviorSFT has been trained with one epoch using the \texttt{adamw\_torch} optimizer ($\beta_1{=}0.9$, $\beta_2{=}0.95$, $\epsilon{=}10^{-8}$). The peak learning rate is $1\times10^{-4}$, decayed with a cosine schedule after a 5\,\% warm-up.  
Training runs in \texttt{bfloat16} on 4×H200 GPUs with an effective batch size of 64 (per-GPU batch 4, gradient accumulation 4); weight decay is 0.01 and gradients are clipped to a max-norm of 1.0.  
For \textsc{BehaviorSFT} we add the special tokens \texttt{<reactive>} and \texttt{<proactive>} and attach LoRA adapters (rank 8, $\alpha=32$) to all linear layers. The best checkpoint, selected by validation accuracy every 100 steps, is reported.

\section{Clinician-in-the-Loop Evaluation Study}
\label{sec:user_study}

To rigorously evaluate our BehaviorSFT agent and validate the proposed dataset, we conducted a comprehensive user study involving board-certified medical professionals. This study was designed to assess the clinical utility of \methodname and to compare the performance of LLM agents exhibiting distinct behavioral characteristics.

\subsection{Participant Recruitment and Profile}
We recruited three medical doctors and each physician underwent a standardized orientation session to familiarize them with the study objectives, annotation tasks, and the custom-developed user interfaces.

\subsection{Study Design and Procedure}
The study was structured into two principal phases, each targeting specific evaluation objectives:

\begin{enumerate}[label=\textbf{Phase \arabic*:}, wide, labelindent=0pt, itemsep=1ex]
    \item \textbf{Dataset Validation}
    
    In this phase, clinicians were tasked with validating a randomly selected subset of tasks (N=30) from the \methodname. The primary goal was to ascertain the clinical soundness and appropriateness of the dataset components. For each presented task, which included a clinical `Task Context', a specific `Question', and multiple-choice `Options' (as illustrated in Figure~\ref{fig:interface_design2}), clinicians utilized a dedicated evaluation panel (Figure~\ref{fig:interface_design1}). Their evaluation encompassed:
    \begin{itemize}[itemsep=0.5ex]
        \item \textbf{Correctness of Ground Truth:} Verifying the accuracy of the designated correct answer among the provided options.
        \item \textbf{Annotator Confidence:} Rating their confidence in their selected answer on a three-point scale (Low, Moderate, High).
        \item \textbf{Task Proactivity Level Assessment:} Evaluating the inherent proactivity level of the question itself on a continuous scale ranging from 0.0 (Reactive) to 1.0 (Proactive). This aimed to capture the degree to which the question prompted an anticipatory or forward-looking response.
        \item \textbf{Clinical Plausibility:} Determining if the task (question and options combined) was clinically plausible and relevant within the given case context, with options "Yes," "No," or "Unsure."
    \end{itemize}
    To ensure comprehensive understanding, clinicians had access to the broader `Case Context', including a `Case Presentation Summary', the `Full Conversation' transcript leading to the task, and an option to refer to the original medical case for in-depth review (Figure~\ref{fig:interface_design3}).

    \item \textbf{Comparative Agent Behavior Evaluation}

    This phase focused on evaluating the quality and safety of responses generated by three distinct LLM agent archetypes when presented with N=10 clinical tasks from \methodname. The agents included: 
    (1) \textbf{BehaviorSFT}: An agent fine-tuned using our proposed \methodname{} approach.
    (2) \textbf{General SFT}: An agent subjected to general supervised fine-tuning without specific behavioral guidance.
    (3) \textbf{ZS + Explicit Instr.}: An agent operating in a zero-shot setting, guided by explicit instructions on desired behavior.

    For each scenario, clinicians were first presented with the `Question Posed to AI' and the `Task Options' (with the correct answer highlighted for their reference). Subsequently, the responses from the three LLM agents were displayed side-by-side (Figure~\ref{fig:interface_design5}). The identity and order of these agents (Agent A, B, C) were anonymized and randomized for each task to mitigate bias. Using the feedback panel shown in Figure~\ref{fig:interface_design4}, clinicians performed the following evaluations:
    \begin{itemize}[itemsep=0.5ex]
        \item \textbf{Comparative Ranking:} Ranking the three agent responses from best (1st) to worst (3rd) using a drag-and-drop mechanism.
        \item \textbf{Safety Assessment:} Identifying and describing any instances of clinically unsafe information, critical errors, or significant omissions in any of the agent responses.
        \item \textbf{Proactivity/Reactivity Appropriateness:} Rating the appropriateness of each agent's proactivity or reactivity level on a 5-point Likert scale (1: Very Inappropriate, 3: Neutral, 5: Very Appropriate).
    \end{itemize}
\end{enumerate}

\subsection{Interface Design for Annotation Tasks}
\begin{figure*}[h!]
    \centering
    \includegraphics[width=0.8\textwidth]{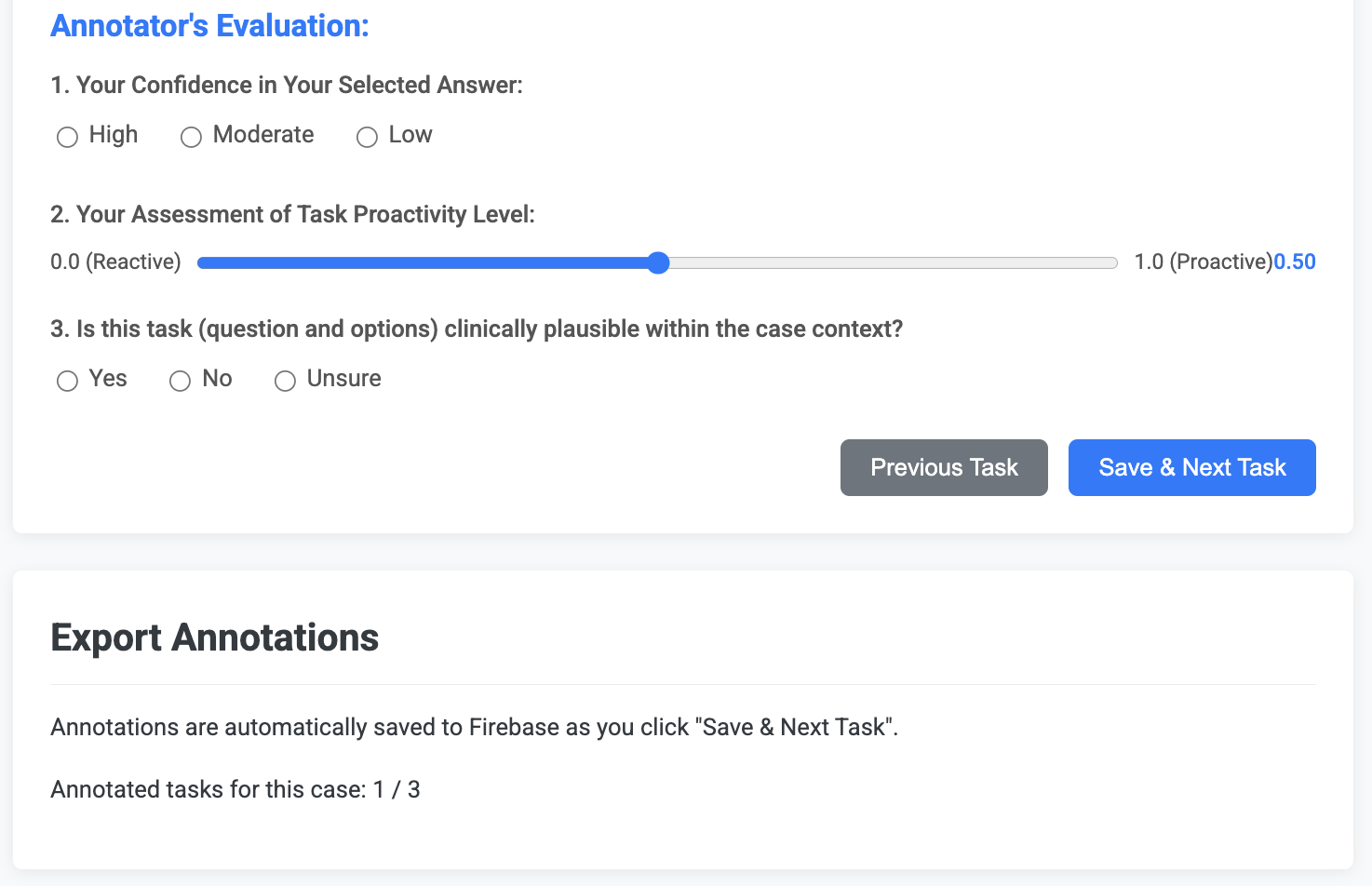} 
    \caption{Interface for \textbf{Dataset Task Validation: Annotator's Evaluation}. Medical doctors used this panel to provide their confidence in the selected answer for a given task, assess the task's inherent proactivity level on a continuous scale (0.0 Reactive to 1.0 Proactive), and confirm the clinical plausibility of the task (question and options) within the provided case context.}
    \label{fig:interface_design1}
\end{figure*}

\begin{figure*}[h!]
    \centering
    \includegraphics[width=0.8\textwidth]{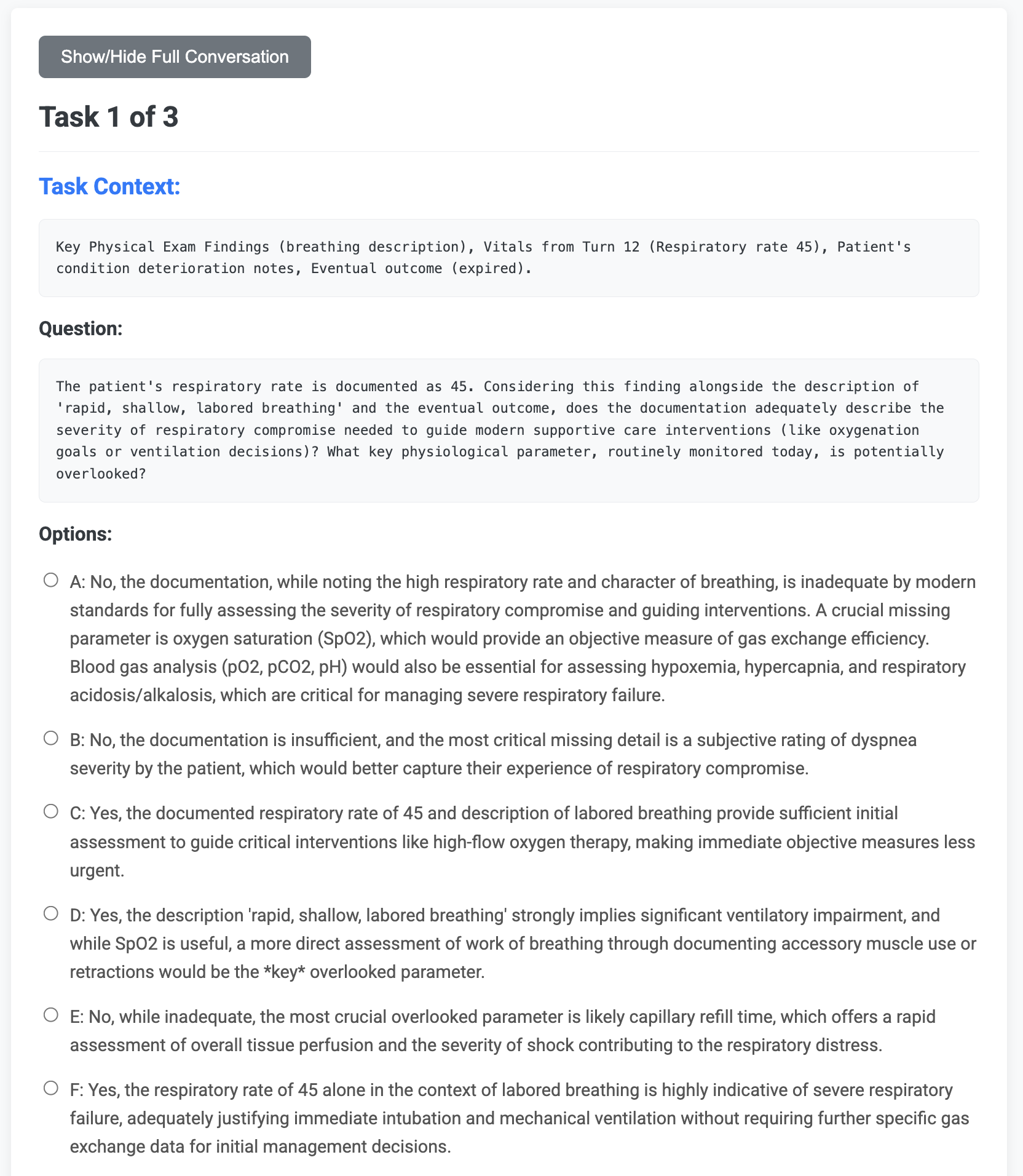} 
    \caption{Interface for \textbf{Dataset Task Validation: Task Presentation}. This view provided clinicians with the `Task Context` (relevant excerpts from the case), the specific `Question` being posed for the \texttt{BehaviorBench} task, and the multiple-choice `Options`, one of which was the ground truth answer they were validating.}
    \label{fig:interface_design2}
\end{figure*}

\begin{figure*}[h!]
    \centering
    \includegraphics[width=0.8\textwidth]{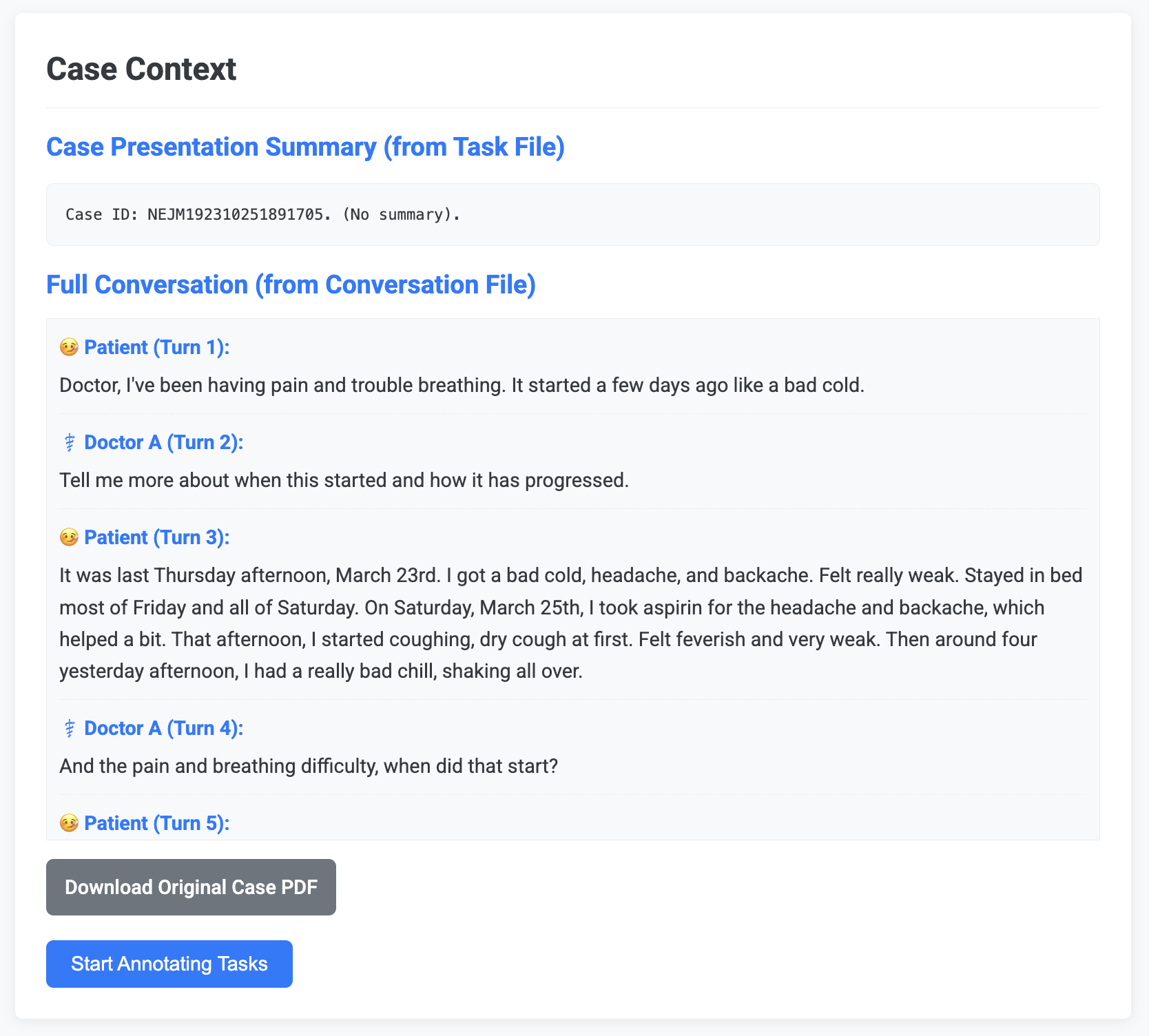} 
    \caption{Interface for \textbf{Dataset Task Validation: Case Context Provision}. To ensure comprehensive understanding, clinicians had access to the broader `Case Context`, including a `Case Presentation Summary` (if available from the task file), the `Full Conversation` transcript leading up to the point of the task, and an option to download the original case PDF for in-depth review.}
    \label{fig:interface_design3}
\end{figure*}

\begin{figure*}[h!]
    \centering
    \includegraphics[width=0.8\textwidth]{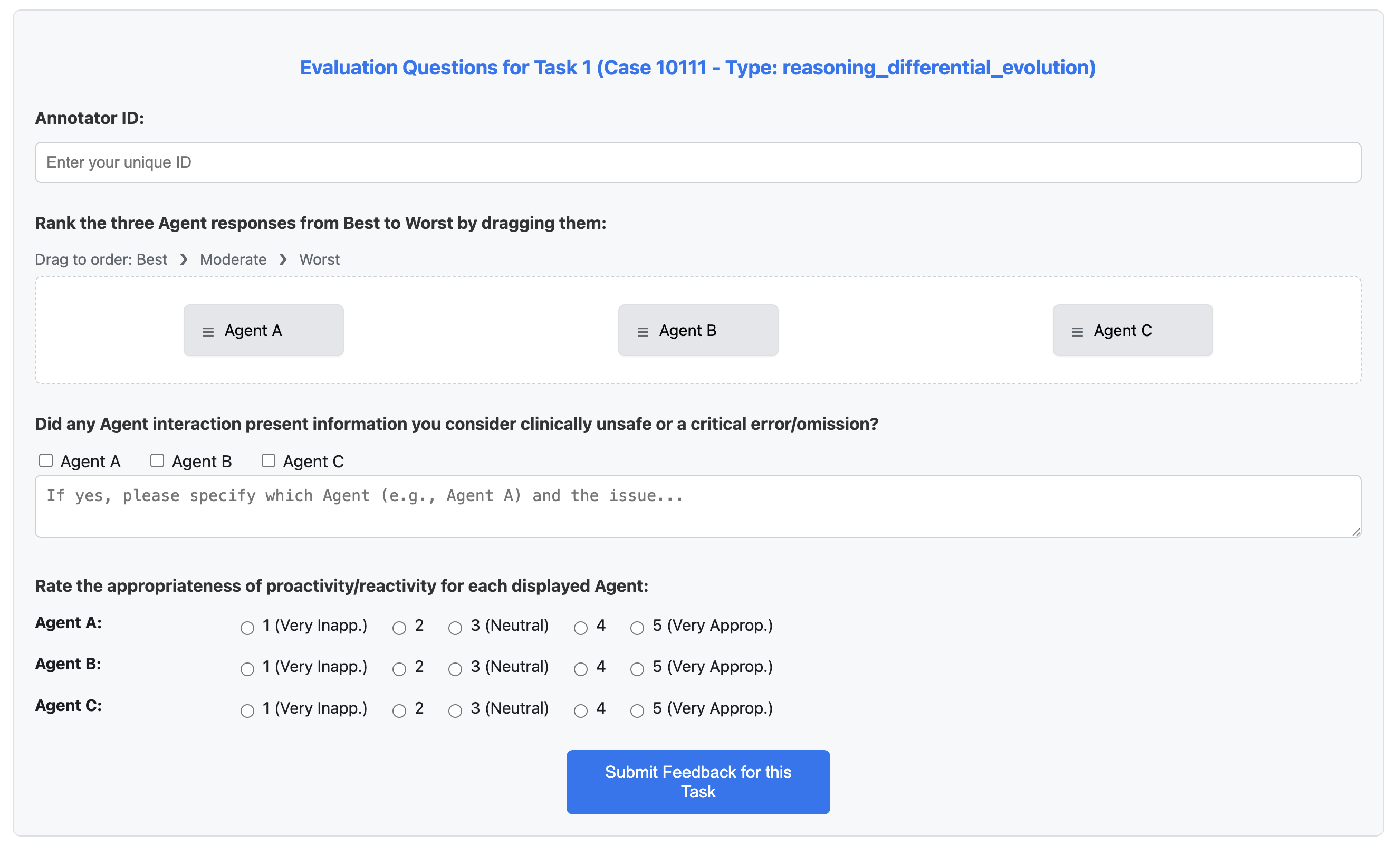} 
    \caption{Interface for \textbf{Agent Behavior Evaluation: Clinician Feedback Panel}. After reviewing the task and agent responses (shown in Figure~\ref{fig:appendix_interface_agent_responses}), medical doctors used this panel to: (1) Rank the three anonymized agent responses (Agent A, B, C) from best to worst via drag-and-drop. (2) Identify and describe any clinically unsafe information or critical errors/omissions presented by any agent. (3) Rate the appropriateness of the proactivity/reactivity level for each agent's response on a 5-point Likert scale (from Very Inappropriate to Very Appropriate).}
    \label{fig:interface_design4}
\end{figure*}

\begin{figure*}[h!]
    \centering
    \includegraphics[width=0.8\textwidth]{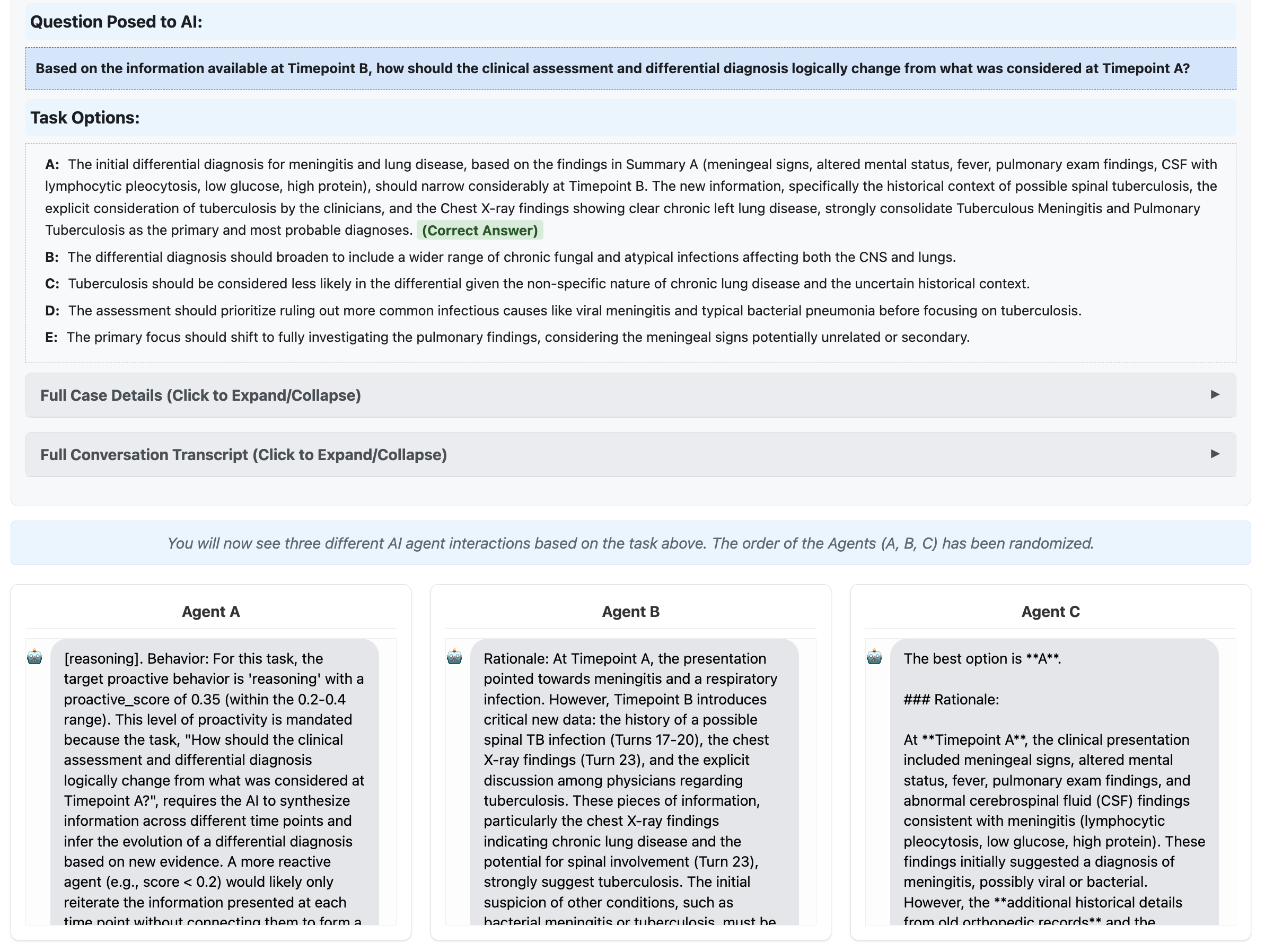} 
    \caption{Interface for \textbf{Agent Behavior Evaluation: Task and Agent Response Display}. For each evaluation scenario, clinicians were presented with the `Question Posed to AI` and the `Task Options` (with the correct answer highlighted for reference). Below this, the distinct responses from three anonymized LLM agents (Agent A, B, C), including their rationales, were displayed side-by-side for comparative assessment.}
    \label{fig:interface_design5}
\end{figure*}

Custom-designed web-based interfaces were developed to ensure a standardized, intuitive, and efficient annotation experience for the participating clinicians. The interfaces were tailored to the specific requirements of each study phase (see Figure \ref{fig:interface_design1}, \ref{fig:interface_design2}, \ref{fig:interface_design3}, \ref{fig:interface_design4} and \ref{fig:interface_design5}).

\subsection{Annotation Results} 

This section presents the quantitative and qualitative findings from the clinician-in-the-loop evaluation study. All reported inter-annotator agreement scores were calculated among the three participating physicians.

\subsubsection{Phase 1: \methodname Validation}
Clinicians evaluated a total of 60 unique tasks from the \methodname. 

\textbf{MCQ Accuracy and Task Plausibility}
The physician annotators demonstrated a high level of accuracy in answering the multiple-choice questions, achieving an overall correctness of \textbf{83.3\%}. This proficiency underscores their expert understanding of the clinical scenarios presented within the dataset.

The clinical plausibility of the tasks was a key validation metric. As shown in Figure~\ref{fig:mcq_plausibility_results}, a substantial majority of tasks (\textbf{80.0\%}) were rated as clinically plausible (``Yes''). No tasks (0.0\%) were rated as definitively ``No'' for plausibility, while 20.0\% were marked as ``Unsure,'' suggesting areas where task framing or context might warrant further refinement or clarification for some annotators.

\textbf{Annotator Confidence Levels}
Annotator confidence in their selected MCQ answers was recorded on a three-point scale. The distribution, illustrated in Figure~\ref{fig:mcq_confidence_results}, reveals that physicians were predominantly ``High'' in their confidence (\textbf{55.0\%}). ``Moderate'' confidence was reported for 36.67\% of answers, while ``Low'' confidence was expressed for only 8.33\% of answers. This general trend towards higher confidence aligns with the observed accuracy.

\textbf{Inter-Annotator Agreement for Dataset Validation}
To ensure the reliability of the dataset validation process, inter-annotator agreement was quantified using the Intraclass Correlation Coefficient (ICC3) for continuous ratings. 

The task proactivity/reactivity slider ratings (0.0-1.0 scale) demonstrated \textit{good} reliability with an ICC3 of \textbf{0.61}. 
This robust agreement scores indicate that the physicians interpreted and applied the validation criteria consistently.

\subsubsection{Phase 2: Comparative Agent Behavior Evaluation Results}
Physicians evaluated agent responses across N=24 unique clinical tasks.
The anonymized agents evaluated were BehaviorSFT, General SFT, and ZS + Explicit Instr.

\begin{figure*}[htbp]
    \centering
    \includegraphics[width=0.9\textwidth]{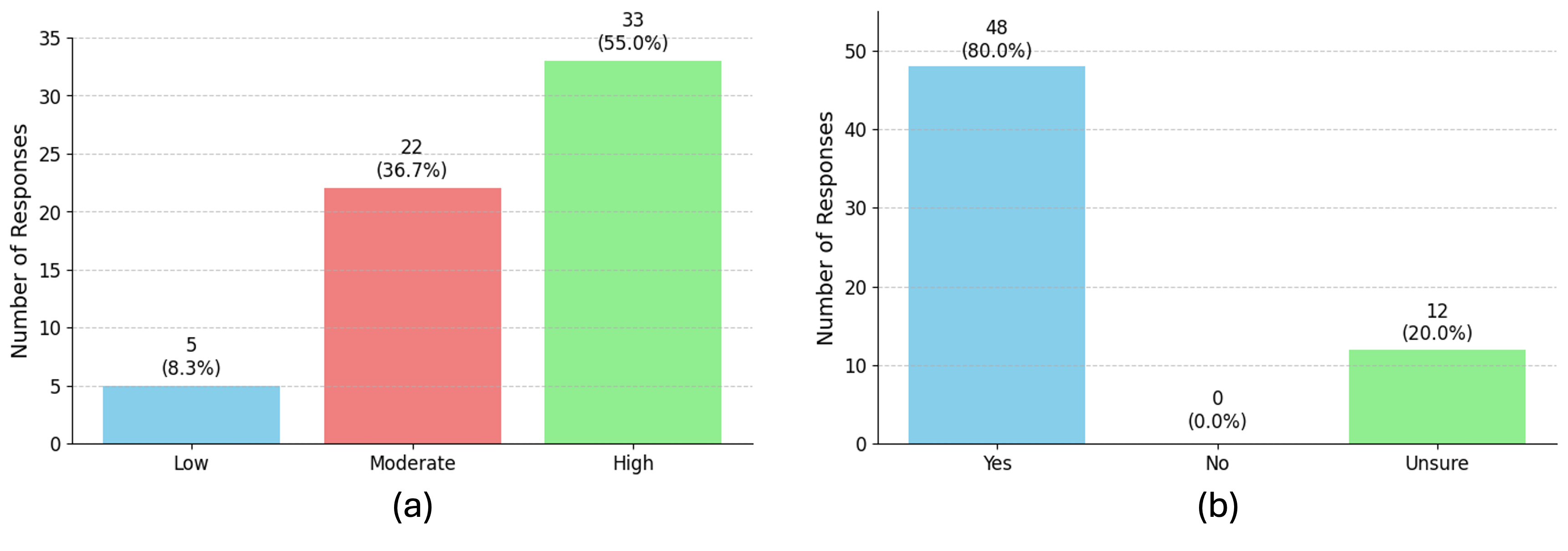}
    \caption{\textbf{(a)} Over half (55.0\%) of the responses were marked as `High' confidence, while `Moderate' confidence accounted for 36.7\%. `Low' confidence was the least frequent category, representing only 8.3\% of responses. \textbf{(b)} The vast majority (80.0\%) of responses affirmed the clinical plausibility (`Yes') of the generated MCQs. A smaller portion (20.0\%) of responses were `Unsure', and no responses found the MCQs implausible (`No'). }
    \label{fig:agent_ranks_results}
\end{figure*}

\begin{figure*}[htbp]
    \centering
    \includegraphics[width=0.9\textwidth]{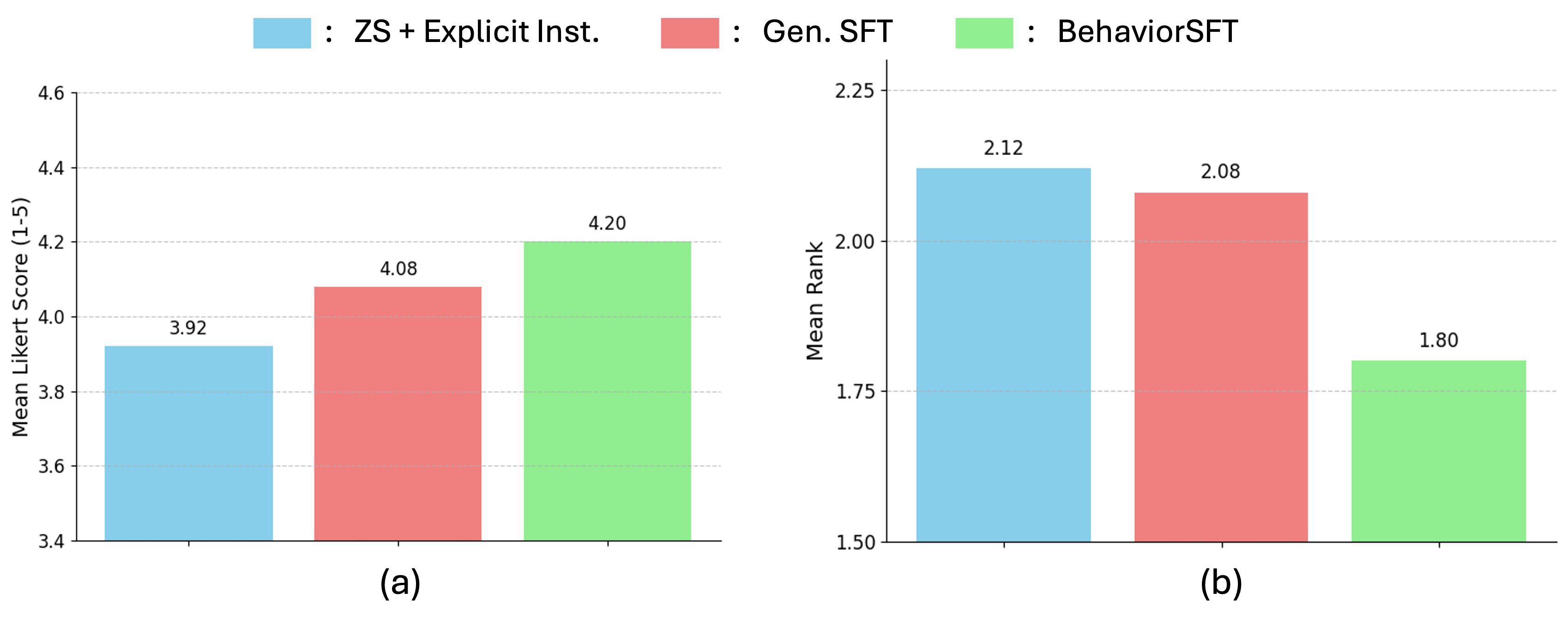}
    \caption{\textbf{(a)} Mean appropriateness scores for agent proactivity/reactivity (5-point Likert scale, higher is better). \textbf{(b)} BehaviorSFT received the lowest (best) mean rank (1.80), suggesting it was most frequently ranked highest by evaluators. Gen. SFT had a mean rank of 2.08, while ZS w/ explicit instruction had the highest (worst) mean rank of 2.12 in a system where lower ranks are better.}
    \label{fig:agent_likert_results}
\end{figure*}

\textbf{Agent Response Ranking and Proactivity/Reactivity Appropriateness}
The primary evaluation involved ranking the three agents. \textbf{Agent A (\methodname{})} received the most favorable rankings, achieving the lowest (best) mean rank of \textbf{1.80} (Figure~\ref{fig:agent_ranks_results}). In terms of the appropriateness of proactivity/reactivity, \textbf{Agent C (ZS + Explicit Instr.)} scored highest with a mean Likert score of \textbf{4.20} out of 5 (Figure~\ref{fig:agent_likert_results}). Agent B (General SFT) had a mean rank of 2.08 and a mean Likert score of 4.08.

\begin{table*}[h!] 
\centering
\footnotesize
\caption{Six-Level Taxonomy for Healthcare AI Agent Autonomy}
\label{tab:autonomy_taxonomy_detailed}
\begin{tabularx}{\textwidth}{>{\bfseries}p{0.7cm} >{\bfseries}p{3.2cm} >{\raggedright\arraybackslash}X >{\raggedright\arraybackslash}X}
\toprule
Level & Name & AI Agent's Role / Capability & Human Clinician's Role \\
\midrule
0 & No Automation & The AI system provides no assistance or automation for any clinical task. & Performs all tasks and makes all decisions related to patient care. The AI system is not involved. \\
\addlinespace
1 & Clinician Assistance & The AI system may provide information, simple alerts based on predefined rules (e.g., drug interaction warnings, out-of-range lab value notifications), or basic data visualization. It does not perform any part of the dynamic clinical task itself. & Performs all dynamic decision-making and actions. Uses the AI as a passive information source or a simple alerting tool. Responsible for interpreting AI-provided information. \\
\addlinespace
2 & Partial Automation (Reactive Support) & The AI system can execute specific, well-defined reactive sub-tasks under direct human supervision based on explicit clinician queries or predefined triggers (e.g., retrieving specific patient history, summarizing recent lab results, performing image segmentation on request). It does not manage the overall clinical situation. & Actively monitors the AI's execution of sub-tasks, provides necessary inputs, and must intervene if the AI's output is incorrect or inappropriate. Responsible for the overall task and integrating AI's contribution. \\
\addlinespace
3 & Conditional Automation (Contextual Proactivity) & The AI system can perform certain proactive tasks and make some decisions within a limited, well-defined clinical context or Operational Design Domain (ODD) (e.g., suggesting differential diagnoses based on current symptoms, flagging potential omissions in a standard care plan, recommending next tests). It can handle some dynamic aspects of the task. & Monitors the AI and the clinical environment. Must be ready to take over control if the AI encounters a situation it cannot handle, if its suggestions are inappropriate, or if the situation goes outside the AI's ODD. \\
\addlinespace
4 & High Automation (Proactive Decision Support) & The AI system can make significant clinical decisions and take proactive actions in most situations within its designed ODD without human oversight for extended periods (e.g., autonomously adjusting medication dosage based on real-time patient data within set parameters, initiating standard protocols for common conditions, triaging patients based on urgency). & Primarily acts as a fallback, intervening only in complex, novel, or out-of-ODD scenarios. Relies on the AI for most routine decisions and actions within the ODD. May oversee multiple AI-managed cases. \\
\addlinespace
5 & Full Automation (Autonomous Operation) & The AI system can perform all clinical tasks and make all decisions that a human healthcare professional can, under all conditions within its defined scope of operation. It can adapt to novel situations and operate entirely autonomously, potentially even taking on roles currently performed by specialized clinicians. & May not be required for tasks within the AI's full operational scope. Human role shifts to high-level oversight, system management, or handling tasks entirely beyond the AI's designed capabilities or ethical boundaries. \\
\bottomrule
\multicolumn{4}{l}{ODD: Operational Design Domain - The specific conditions under which a given AI system or feature is designed to function.} \\
\end{tabularx}
\end{table*}

\end{document}